\documentclass[times,referee,twocolumn,final]{elsarticle}

\usepackage{ycviu}
\usepackage{framed,multirow}

\usepackage{amssymb}
\usepackage{latexsym}

\usepackage{url}
\usepackage{xcolor}
\definecolor{newcolor}{rgb}{.8,.349,.1}

\usepackage{epsfig}
\usepackage{booktabs}
\usepackage{xcolor}
\usepackage{mathtools}
\usepackage{multirow}
\usepackage{array}
\usepackage{float}
\usepackage{nicefrac}    
\usepackage{amsfonts} 
\usepackage{dsfont}
\usepackage{flushend}

\usepackage{amsmath}
\usepackage{amssymb}
\usepackage{dsfont}
\usepackage{bm}
\usepackage{xcolor,soul,colortbl}
\usepackage{array}
\usepackage{multirow}
\usepackage{subcaption}
\usepackage{dashrule}
\usepackage{microtype}
\usepackage{enumitem}
\usepackage{breakcites}

\usepackage{lipsum}
\usepackage{mdframed}
\definecolor{real_feat}{RGB}{127,0,255}
\definecolor{buda_feat}{RGB}{240,230,140}
\definecolor{baseline_feat}{RGB}{150,75,0}
\definecolor{zs3_feat}{RGB}{51,51, 255}

\definecolor{col_tuk}{RGB}{255,205,50}
\definecolor{col_car}{RGB}{0,0,143}
\definecolor{col_truck}{RGB}{0,0,68}
\definecolor{col_tree}{RGB}{106,143,28}
\definecolor{col_road}{RGB}{129,61,128}
\definecolor{col_sidewalk}{RGB}{243,30,233}
\definecolor{col_terrain}{RGB}{152,250,152}
\definecolor{col_person}{RGB}{253,0,0}
\definecolor{col_bus}{RGB}{0,57,100}
\definecolor{col_train}{RGB}{0,79,99}
\definecolor{col_moto}{RGB}{0,0,231}
\definecolor{col_sky}{RGB}{0,131,181}
\definecolor{col_building}{RGB}{68,68,68}
\definecolor{col_animal}{RGB}{249,196,148}

\definecolor{col_truck_coco}{RGB}{129,61,128}
\definecolor{col_person_coco}{RGB}{193,129,129}
\definecolor{col_background_coco}{RGB}{0,0,0}
\definecolor{col_tv_coco}{RGB}{0,64,128}
\definecolor{col_sofa_coco}{RGB}{1,192,0}
\definecolor{col_cow_coco}{RGB}{57,125,0}
\definecolor{col_cat_coco}{RGB}{62,0,0}
\definecolor{col_laptop_coco}{RGB}{193,62,0}
\definecolor{col_giraffe_coco}{RGB}{64,64,2}
\definecolor{col_zebra_coco}{RGB}{122,196,127}
\definecolor{col_bench_coco}{RGB}{0,192,127}
\definecolor{col_chair_coco}{RGB}{194,0,1}
\definecolor{col_sheep_coco}{RGB}{128,64,3}

\newcommand{\mbf}[1]{\ensuremath{\mathbf{#1}}} 
\newcommand{\mbs}[1]{\ensuremath{\boldsymbol{#1}}} 
\newcommand{\mc}[1]{\ensuremath{\mathcal{#1}}}

\newcommand{\ba}{\mbf{a}}

\newcommand{\bz}{\mbf{z}}

\newcommand{\cD}{\mc{D}}

\newcommand{\mm}[1]{\ensuremath{\bm{#1}}} 

\newcolumntype{C}[1]{>{\centering\arraybackslash}p{#1}}	
\newcolumntype{L}[1]{>{\raggedright\arraybackslash}p{#1}}
\newcolumntype{R}[1]{>{\raggedleft\arraybackslash}p{#1}}

\newcommand{\src}[1]{#1^s}
\newcommand{\trg}[1]{#1^t}
\newcommand{\fakefeat}{\hat{\mbs{f}}}
\newcommand{\realfeat}{\mbs{f}}
\newsavebox\CBox
\def\textBF#1{\sbox\CBox{#1}\resizebox{\wd\CBox}{\ht\CBox}{\textbf{#1}}}

\usepackage[pagebackref=true,breaklinks=true,letterpaper=true,colorlinks,bookmarks=false]{hyperref}

\journal{Computer Vision and Image Understanding}

\begin{document}

\ifpreprint
  \setcounter{page}{1}
\else
  \setcounter{page}{1}
\fi

\begin{frontmatter}
\title{Handling new target classes in semantic segmentation with domain adaptation}

\author[1]{Maxime \snm{Bucher}} 
\cortext[cor1]{Paper under review }
\author[1]{Tuan-Hung \snm{Vu}}
\author[1,2]{Matthieu \snm{Cord}}
\author[1]{Patrick \snm{Pérez}}

\address[1]{Valeo.ai, Paris, France}
\address[2]{Sorbonne University, Paris, France}

\begin{abstract} 
In this work, we define and address a novel domain adaptation (DA) problem in semantic scene segmentation, where the target domain not only exhibits a data distribution shift w.r.t. the source domain, but also includes novel classes that do not exist in the latter. 
Different to ``open-set''~\cite{panareda2017open} and ``universal domain adaptation''~\cite{you2019universal}, which both regard all objects from new classes as ``unknown'', we aim at explicit test-time prediction for these new 
classes.
To reach this goal, we propose a framework that leverages domain adaptation and zero-shot learning techniques to enable ``boundless'' adaptation in the target domain. It relies on a novel architecture, along with a dedicated learning scheme, to bridge the source-target domain gap while learning how to map 
new classes'~\textit{labels} to relevant visual representations.
The performance is further improved using self-training on target-domain pseudo-labels.
For validation, we consider different domain adaptation set-ups, namely synthetic-2-real, country-2-country and dataset-2-dataset.
Our framework outperforms the baselines by significant margins, setting competitive standards on all benchmarks for the new task.

Code and models are available at:~\url{https://github.com/valeoai/buda}.
\end{abstract}
\end{frontmatter}

\section{Introduction}
\label{sec:intro}
\begin{figure*}[t]
\captionsetup[subfigure]{labelformat=empty}
	\begin{center}
		\begin{subfigure}[t]{0.32\textwidth}\centering
			\caption{\small \textbf{Input image}}\vspace{-0.2cm}
		\includegraphics[trim=0cm 0.0cm 0cm 0.1cm, clip=true,width=1.\textwidth]{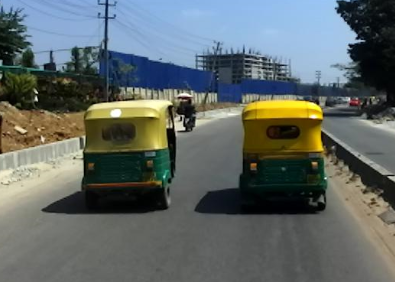}
		\end{subfigure}
		\begin{subfigure}[t]{0.32\textwidth}\centering
			\caption{\small \textbf{Closed-set UDA}}\vspace{-0.2cm}
			\includegraphics[trim=0cm 0.2cm 0cm 0.1cm, clip=true,width=1.\textwidth]{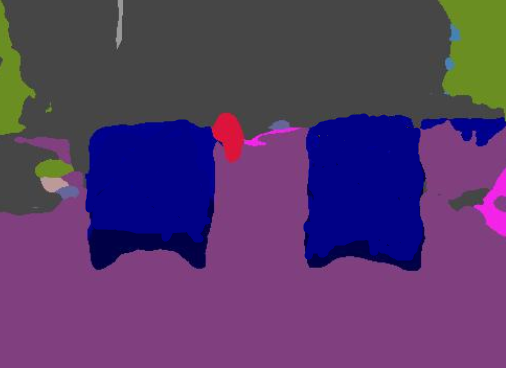}
		\end{subfigure}
		\begin{subfigure}[t]{0.32\textwidth}\centering
			\caption{\small \textbf{BudaNet}}\vspace{-0.2cm}
			\includegraphics[trim=0cm 0.2cm 0cm 0.0cm, clip=true,width=1.\textwidth]{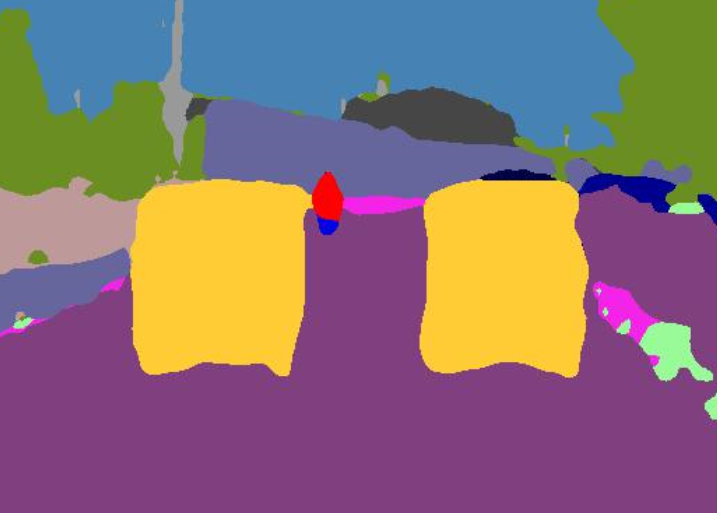}
		\end{subfigure}\\
	\end{center}
	\caption{\textbf{Illustration of BudaNet, the proposed approach to the new problem of boundless UDA for semantic segmentation}. At test time, for an input image (left), we display the closed-set UDA segmentation result (middle) as well as BudaNet's segmentation result (right). 
			Both ``\setlength{\fboxsep}{1pt}\colorbox{col_tuk}{\textcolor{white}{tuk-tuk}}'' vehicles, which are from a new class only appearing in the target domain, have been correctly identified by BudaNet: Our approach is able to deal with new classes for which no annotated images have been provided during training. The model trained following the closed-set UDA setting wrongly predicts these new vehicles as a spatial mix of \setlength{\fboxsep}{1pt}\colorbox{col_car}{\textcolor{white}{car}} and	\setlength{\fboxsep}{1pt}\colorbox{col_truck}{\textcolor{white}{truck}}.}
	\label{fig:teaser}
\end{figure*}

Detailed interpretation of operating environments is crucial for autonomous systems like self-driving cars or delivery droids.
Aiming at such an exhaustive scene understanding, most recent vision systems conduct semantic segmentation, the task of predicting semantic classes for all scene pixels.
As for other perception modules, significant shifts between train (source domain) and test (target domain) distributions drastically degrade the segmentation performance.
Many works, \textit{e.g.}, \cite{hoffman2018cycada,tsai2018learning,vu2019advent}, propose unsupervised domain adaptation (UDA) techniques to address such \emph{domain gaps}, while alleviating the taxing need for data labeling in the target domain. Besides these domain adaptation (DA) works, where the same categories are assumed in both domains, 
few recent works consider more general set-ups in which source and target label sets differ.
For example, partial DA~\cite{cao2018partial,zhang2018importance} is the case where the source label set contains the target one.
Differently, open-set DA~\cite{panareda2017open} assumes that each domain holds a set of ``private'' classes besides the ones they share.
More challenging, universal DA~\cite{you2019universal} considers having a completely unknown target label set.
These preceding efforts contribute all to make domain adaptation closer to real-world applications. 
In open-set and universal DA settings, objects from that classes that are absent in the source domain  
are all classified as ``unknown''. 
Nevertheless, for real-life scenarios where the target label set is indeed \emph{boundless}, one could expect the final system to predict new classes explicitly.
For example, one might require that the perception model, once trained on European driving scenes, 
behaves well on Indian streets where vehicles similar to European ones cohabit with specific ones like ``tuk-tuk''. 
Zhuo \textit{et al.}~\cite{zhuo2019unsupervised} address this realistic set-up for classification and propose a unified framework to handle new classes at test time while minding the domain gap.

All the above-mentioned DA works take image classification as the only test-bed.

Different to those works, we address here semantic segmentation with UDA, \textit{i.e.}, the task of exhaustively classifying all pixels in input images in presence of a domain gap.  We aim at pixel-wise recognition of objects from both shared and new classes in the target domain but, as each image may contain multiple objects, this UDA task is much more challenging than classification. As required in several real-life applications, the detailed analysis of urban scenes is an example where such complex spatial arrangements of multiple object categories are common place.     

In this segmentation set-up, pixels of shared and new classes can co-occur in the same scene, requiring dedicated strategies to carefully handle source-target domain alignments; for example, as new-class objects only exist in the target domain, a naive alignment might undesirably force their visual features close to those from shared-class objects.

We call this framework ``boundless'' unsupervised domain adaptation'', BUDA in short, and we propose ways to attack it in the present work.
In particular, we introduce a full semantic segmentation pipeline, BudaNet, that jointly addresses two main challenges: (1) Bridging the gap between source and target domains; (2) Learning discriminant visual representations for new-class objects in the target domain. 
Within a standard semantic segmentation framework, we propose a UDA strategy to mitigate the distribution gap for shared classes across domains, without causing unwanted misalignment to new ones.
We then introduce a novel domain-aware model to generate, for all classes, pixel-level visual features for both domains.
By using this generative model, we collect features from the new target-domain classes, along with features from shared classes in the source domain, to learn the last classification layer of BudaNet.
Last, we refine BudaNet with a step of self-training using pseudo-labels on the target domain.
The resulting behavior of our system is illustrated in Figure~\ref{fig:teaser}.

With this new form of DA task and the proposed BudaNet approach to solve it, we advocate for more practical and ubiquitous domain adaptation in which the target label set is boundless.
\begin{figure*}[]
\centering
\includegraphics[scale=0.5]{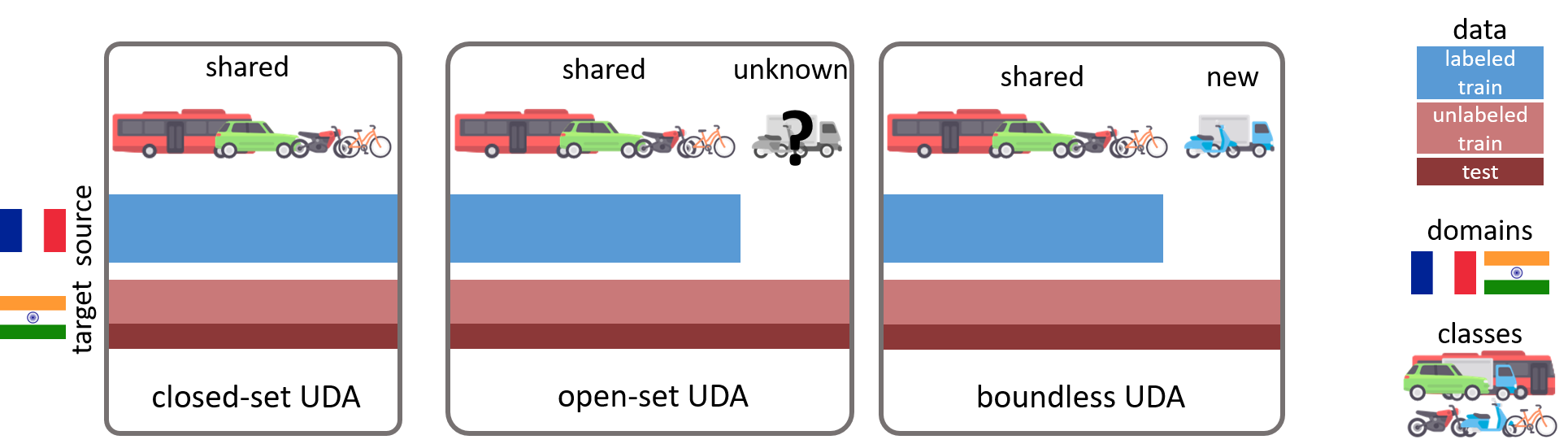}
\caption{\textbf{Different UDA settings}. In contrast with classic UDA (``closed-set'' hypothesis), boundless UDA (BUDA) assumes that source and target label sets differ. In addition, unlike the ``open-set'' setting where objects from the private target classes are classified in a single \textit{unknown} category, BUDA allows the target classes to be explicitly predicted. 
}
\label{fig:buda_diagram}
\end{figure*}

\section{Related work and positioning}
In this section, we briefly go through previous UDA works and position our setting with respect to them. We also review some zero-shot learning techniques that play an important role in the proposed framework.

\vspace{-0.2cm}
\paragraph{Unsupervised domain adaptation} 
Most existing UDA works consider the classic setting in which source and target label sets are the same.
Though approaching UDA via different angles, these works are in the same vein of learning task-dependent domain-invariant features, \textit{i.e.}, minimizing inter-domain discrepancy of feature distributions. Popular techniques include regularizing the maximum mean discrepancy~\cite{long2015learning}, matching deep activation correlation~\cite{sun2016deep}, aligning source-target distributions via adversarial training~\cite{ganin-icml2015,hoffman2018cycada}, self-training~\cite{li-cvpr2019,xie2018learning,zou2018domain} or self-ensembling~\cite{french2017self,zhu2005semi}.
Recently, UDA for complex recognition tasks like detection~\cite{chen2018domain} and semantic segmentation~\cite{hoffman-arxiv2016} have received more attention.
Such tasks often require special techniques to handle as well the spatial layout~\cite{tsai2018learning,vu2019advent} or class proportions~\cite{zou2018domain}.
While the standard UDA setting facilitates investigations and helps gain fundamental insights, it is still far from the real-life scenarios.

Some recent works propose techniques in more relaxed UDA settings.
In those cases, differences between source and target label sets make approaches by direct distribution alignment less effective.
For partial DA, Cao~\textit{et al.} \cite{cao2018partial} align class-wise distributions using multiple domain discriminators, while Zhang~\textit{et al.} \cite{zhang2018importance} adopt an auxiliary domain classifier to estimate source-domain sample importance.
In open-set DA, where source and target domains can hold private label sets, a common practice is to use an ``unknown'' class gathering all target domain's private objects~\cite{panareda2017open,saito2018open}.
In~\cite{you2019universal}, the authors introduce a more extreme setting coined ``universal'' DA, which imposes no prior knowledge on the target label set.

\vspace{-0.2cm}
\paragraph{Dealing with news classes} 
Zero-Shot Learning (ZSL) aims to recognize new classes based solely on their semantic associations with previously seen classes. To this end, attributes \cite{lampert2013attribute}, description \cite{reed2016learning} or even word embedding \cite{frome2013devise} have been shown to be shared semantic representations that allow transferring knowledge from seen to unseen classes. In this work, we use word2vec~\cite{mikolov2013distributed} as it is extracted at minimal cost and it alleviates the need to define and assign hundreds of attributes.

ZSL for image classification has been actively studied in the literature~\cite{akata2015evaluation,bucher2016improving,bucher2017generating,gao2018joint,kodirov2017semantic,romera2015embarrassingly,socher2013zero,xian2018feature,zhang2015zero} and existing methods can be generally categorized into the following two groups.  
The first group \cite{akata2015evaluation,bucher2016improving,kodirov2017semantic,romera2015embarrassingly,socher2013zero,zhang2015zero} addresses ZSL as an embedding problem and maps image data and class descriptions into a common space where semantic similarity translates into spacial proximity. 
The second group \cite{bucher2017generating,gao2018joint,xian2018feature} of methods learns from visual and semantic features of seen classes and produces generators that can  synthesize visual features based on class semantic descriptions. The synthetic features are then used to train a standard classifier for object recognition. This second type of methods has been shown to be more effective than embedding approaches as it reduces the inherent bias toward seen classes. In this work, we leverage this type of generative techniques.

More and more ZSL works~\cite{zou2018domain,kumar2018generalized,schonfeld2019generalized} address the generalized setting (GZSL) where both seen and unseen classes are jointly evaluated at test time.
From a practical point of view, GZSL's evaluation protocol better reflects the real-life performance of zero-shot models.

Very recently, generalized zero-shot learning has been extended to semantic segmentation \cite{bucher2019zero,kato2019zero,xian2019semantic}. Bucher \textit{et al}.\,\cite{bucher2019zero} introduce the task by combining a deep visual segmentation model with an approach to generate visual representations from semantic word embeddings. Kato \textit{et al}.\,\cite{kato2019zero} propose a variational mapping of the class labels' embedding vectors to the visual space.
Xian \textit{et al}.\,\cite{xian2019semantic} introduce a model with a visual-semantic embedding module to encode images in the word embedding space and a semantic projection layer that produces class probabilities.
In this work, we evaluate our models using the GZSL evaluation protocol for semantic segmentation proposed in~\cite{bucher2019zero}.

\paragraph{Transductive zero-shot learning}
Transductive ZSL solves ZSL in a semi-supervised learning manner, \textit{i.e.}, all data including test ones are available at train time and, as with classic ZSL, there is no domain gap between base-class labelled samples and new-class unlabelled test ones. The latter can be utilized to form clearer decision boundaries for both base and new classes. 
Rohrbach \textit{et al}.\,\cite{rohrbach2013transfer} explore the manifold structure of new classes by graph-based label propagation. Fu \textit{et al}.\,\cite{fu2015transductive} extend the label propagation with a multi-view hyper-graph. Several approaches adopt a joint learning framework to train on labeled and unlabeled data separately \cite{guo2016transductive, kodirov2015unsupervised, yu2018transductive}. Such a training can be in the semantic space \cite{guo2016transductive}, the visual space \cite{kodirov2015unsupervised} or a latent space \cite{yu2018transductive}. Other efforts attempt to refine visual-semantic embeddings iteratively with unlabeled unseen data \cite{yu2017transductive}. A domain-invariant projection is learnt in \cite{zhao2018domain}, which maps visual features to semantic embeddings and then reconstructs back the same visual features. Recently, \cite{song2018transductive} described a transductive unbiased embedding to improve generalized ZSL performance.

\paragraph{Positioning}

In this paper, we consider for semantic segmentation the ``open UDA" setting described in~\cite{zhuo2019unsupervised} for classification.
In this setting, the system is expected to handle explicitly objects from new classes in the target domain. Here, the target label set is thus unbounded.
In Figure~\ref{fig:buda_diagram}, we illustrate this boundless UDA setting alongside other UDA settings.
Opposite to the open-set setting in~\cite{saito2018open}, the source label set in BUDA is a subset of the target label set. BUDA is different from both open-set and universal DA as it explicitly requires prediction for all labels at test time.
Also, as briefly mentioned in Section~\ref{sec:intro} and later elaborated in Section~\ref{sec:budanet}, the natural co-occurrence of objects from both shared and private classes in train and test target-domain images makes BUDA more challenging.

The BUDA framework differs from classic ZSL since training images with new classes are available but are unlabeled, whereas they would not be available at all during training in a pure ZSL setting.
Our scenario is also different from transductive learning as: (1) training and test sets of target-domain samples are disjoint and (2) test images stem from a domain that differs from the labeled training domain, opening the door to the domain-shift problem.

BUDA raises jointly zero-shot problems (handling private classes) and domain adaptation ones (mitigating domain shifts); traditional ZSL and DA approaches are thus insufficient here.
One the one hand, a direct application of zero-shot techniques fails to close the distribution gaps between source and target shared classes.
As one must leverage shared-class features and semantic relations between shared and private classes to ``generate'' private-class features, source/target misalignment of the shared-class features results in an erroneous mapping to private-class features in the target domain.
On the other hand, a direct distribution alignment across domains should be avoided to not undesirably mix-up features of shared and private classes.
These two points are discussed in Sections~\ref{sec:budanet_shared} and~\ref{sec:budanet_private}.
\section{Proposed framework: BudaNet}
\label{sec:budanet}

We first formulate the new task of semantic segmentation with boundless UDA. 
Given a set $\mathcal{C}_s$ of class labels in the source domain,
a training set $\src{\cD}$ is available in this domain. It is composed of pairs $(\src{\mm x}, \src{\mm y})$ formed by a color image $\src{\mm x}\in \mathbb{R}^{H\times W \times 3}$ of size $H\times W$ and the associated $C_s-$class ground-truth one-hot segmentation map $\src{\mm y} \in \{0,1\}^{H\times W\times C_s}$, where $C_s = |\mathcal{C}_s|$. In addition, a set $\trg{\cD}$ of unlabelled color images from the target domain is also available at train time.

Aiming for a more practical DA, we argue that, given a minimal semantic prior, the final system should be capable of explicitly predicting new classes, \textit{i.e}, which are not part of $\mathcal{C}_s$, hence with no instances visible in the annotated source-domain images.
In search of such minimal and practically accessible prior, 
we opted for the \textit{names} of these new classes. These names are assumed to be known beforehand, certainly something effortless to achieve. 
In order to exploit the semantic relations between the classes of interest, we adopt the `word2vec' model~\cite{mikolov2013distributed} learned on a dump of the Wikipedia corpus.

In BUDA, the set $\mathcal{C}_s$ of classes in the source domain is fully shared with the target domain -- the source domain does not hold privates classes --, while the latter hold a private set of new classes we denote $\mathcal{C}_p$ and $C_p$ its size. 
At run-time, we want to classify each pixel of target-domain images as one of the categories in $\mathcal{C}_s \cup \mathcal{C}_p$.
In BUDA, we know beforehand all these labels.

\begin{figure*}[t!]
 \centering
 \includegraphics[scale=0.9]{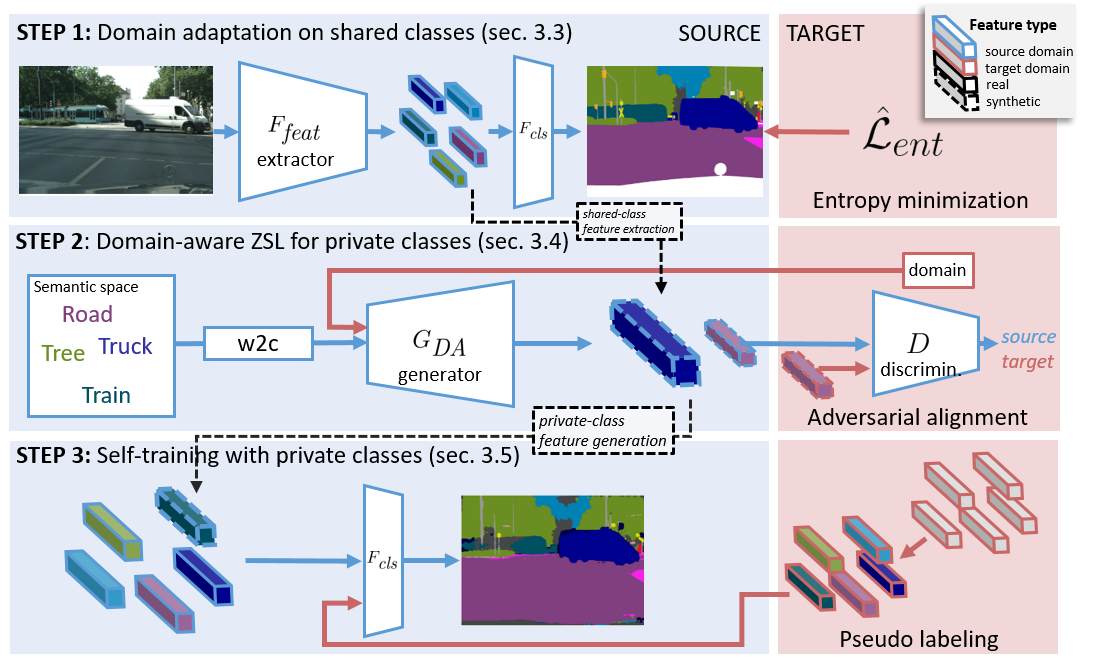}
 \caption{\textbf{BudaNet's architecture and learning framework.} Our BudaNet learning strategy consists of  three steps: Train the semantic segmentation network with domain alignment only for shared classes; Extract image features and use them as supervision for domain-aware generative model training; Combine the generated features with real ones to fine-tune the classification layer. The red-background pane (right) corresponds to the three proposed strategies for domain alignment, zero-shot learning on target domain and self-training with private classes. Black-dash arrows indicate connections between the steps. Colored arrows distinguish source-domain from target-domain flows.}
 \label{fig:gc_gen}
 \end{figure*}

\subsection{Introduction to our strategy}

In this work, we propose a multi-step framework, coined as BudaNet, with dedicated strategies to address the aforementioned concerns.
We start from an existing semantic segmentation model trained with a supervised loss on source-domain data and an unsupervised loss on target-domain train data (Step 1 in Figure\,\ref{fig:gc_gen}). The unsupervised criterion mitigates the source-target distribution gap over the shared classes without causing unwanted misalignment to the private ones. 

Once trained, this first semantic segmentation model is limited to trained categories and, hence, unable to recognize new classes.
To allow the model to recognize both shared and private categories, we propose to generate synthetic training data for private classes. 
This is obtained with a generative model conditioned on the semantic representation of these classes. The generator's outputs attempt to mimic the visual features computed by the current semantic segmentation model (Step 2 -- blue part in Figure\,\ref{fig:gc_gen}).
As no existing UDA technique guarantees perfect and universal alignment between domains, we decided to take domain information into account when training the generator. This is done by conditioning not only on the class embedding but also on the domain.
In addition, we introduce an additional adversarial discriminator that tries to distinguish source-target generated features (Step 2 -- red part in Figure\,\ref{fig:gc_gen}). 

Once the generator is trained, many synthetic features can be produced for private classes, and combined with real samples from shared classes. This new set of training data is used to retrain the classifier of the segmentation network so that it can now handle both shared and private classes (Step 3 -- blue part in Figure~\ref{fig:gc_gen}). The classifier is used to extract pseudo-labels on target-domain training set. 
The whole segmentation network is then fine-tuned again with these pseudo-labels (Step 3 -- red part in Figure\,\ref{fig:gc_gen}).

Section~\ref{sec:based_pipeline} overviews the base zero-shot pipeline which helps to handle never-seen classes in semantic segmentation.
Section~\ref{sec:budanet_shared} introduces our UDA strategy to align shared classes across domains with minimal negative alignment effects on private-class features.
In Section~\ref{sec:budanet_private}, we present the novel domain-aware generative model to synthesize private-class features for the target domain.
Lastly, Section~\ref{sec:budanet_cls} details the final self-training procedure for the classifier.

\subsection{Base zero-shot pipeline}
\label{sec:based_pipeline}

We now revisit the pure adaptation-free zero-shot semantic segmentation (ZS3) pipeline from~\cite{bucher2019zero}, trained only on source-domain images in our context.
This pipeline is built on top of an existing semantic segmentation network $F$, \textit{i.e.}, DeepLabv3+~\cite{chen2018encoder}.
To facilitate explanations, we decompose $F$ into two consecutive parts: $F_{\textit{feat}}$ as the feature extractor, which outputs pixel-wise features $\boldsymbol{f}\in\mathbb{R}^{d_f}$, and $F_{\textit{cls}}$ as the final $1 \times 1$ convolutional classification layer which associates class-scores to each pixel-wise feature (its output size thus depends on the number of considered classes). 

The base zero-shot pipeline consists of three steps:
\begin{enumerate}
	\item Training of a segmentation network $F$ using source-domain supervision on shared classes. Once trained, the feature extractor $F_{\textit{feat}}$ can extract visual features from shared-class objects in source-domain images. 
	
	\item Training of a generative network $G$, conditioned on shared-class embeddings, to generate corresponding source-domain visual features. The ground-truths are the shared-class features extracted after Step 1. The high-level idea is that geometric relations among classes in the embedding space are transferred to the generated feature space, which will help $G$ to handle private classes.
	
	\item Training of the last classification layer $F_{\textit{cls}}$ of $F$, now for both shared and private classes, with private-class features (generated by $G$ after Step 2) and shared-class features (computed by $F_{\textit{feat}}$ after Step 1).
\end{enumerate}
The final model is composed of $F_{\textit{feat}}$ from Step 1 and $F_{\textit{cls}}$ from Step 3.
In detail, the segmentation network $F$ and the classifier $F_{\textit{cls}}$ are trained using a standard segmentation cross-entropy loss $\mathcal{L}_{\textit{seg}}$.
Given a source sample $(\src{\mm x}, \src{\mm y})$ 
and ${\mm P}^{\src{\mm x}}$ its soft-segmentation map predicted by $F(\src{\mm x})$,
this loss reads: 
    \begin{equation}
        \mathcal{L}_{\textit{seg}}(\src{\mm x}, \src{\mm y}) = - \sum_{h=1}^H\sum_{w=1}^W\sum_{c\in\mathcal{C}} \src{\mm y}[h,w,c] \log \mm P^{\src{\mm x}}[h,w,c].
    \end{equation}
    
For the training of $F$ in Step 1 (resp. in Step 3), we set $\mathcal{C}$ to $\mathcal{C}_s$ (resp. $\mathcal{C}_s\cup\mathcal{C}_p$). 
We follow~\cite{bucher2019zero} and adopt a generative moment-matching network (GMMN) as $G$, trained with a maximum mean discrepancy loss~$\mathcal{L}_{\textit{GMMN}}$.
Given a random sample $\bz\in\mathbb{R}^{d_z}$ from a fixed multivariate Gaussian distribution, the semantic embedding $\ba\in\mathbb{R}^{d_a}$ and the domain indicator $d$ (set as 1 for source and 0 for target), new pixel-level  feature sets are generated as $\fakefeat = G(\ba, d, \bz;\theta_{G})$, where $\theta_{G}$ are the parameters of $G$. With two random populations $\mathcal{F}(\ba, d)$ of real features and $\widehat{\mc{F}}(\ba, d;\theta_{G})$ of generated ones, we have: 
    \begin{align}
        \mc{L}_{\textit{GMMN}}(\ba, d) = &  \sum_{\realfeat,\realfeat'\in\mc{F}(\ba,d)} k(\realfeat,\realfeat')  + \sum_{\fakefeat,\fakefeat'\in\widehat{\mc{F}}(\ba,d;\theta_{G}) } k(\fakefeat,\fakefeat') \nonumber\\ & - 2\sum_{\realfeat\in\mc{F}(\ba,d)} \sum_{\fakefeat\in\widehat{\mc{F}}(\ba,d;\theta_{G})} k(\realfeat,\fakefeat),
    \end{align}
    where $k$ is the Gaussian kernel, $k(\realfeat, \realfeat') = \exp(-\frac{1}{2\sigma^2} \|\realfeat-\realfeat' \|^{2})$ with bandwidth parameter $\sigma$.

Figure~\ref{fig:gc_gen}, to the left, illustrates the ZS3 pipeline.
Our proposed BudaNet is built upon such a base three-steps paradigm.

\subsection{BudaNet's Step~$1$: Domain adaptation on shared classes}
\label{sec:budanet_shared}
A pure zero-shot model like ZS3 is able to handle simultaneously base and new classes from the same domain but 
fails to address the domain gap exhibited in BUDA.
One straightforward solution for BUDA is to use UDA techniques while training $F$ in Step 1, with the hope that $F_{\textit{feat}}$ can extract domain-invariant features.
Additional unlabeled target-domain images are used for this purpose.

To this end, we study two recent UDA techniques in segmentation, MinEnt (self-training) and AdvEnt (adversarial training), both introduced in~\cite{vu2019advent}.
With MinEnt, the unsupervised entropy loss $\mathcal{L}_{\textit{ent}}$, applied on target-domain samples, is jointly optimized with the supervised segmentation loss $\mathcal{L}_{seg}$ on source-domain samples.
This loss is the sum of pixel-wise predictive entropies in target domain.
Given a target sample $\trg{\mm x}$ with predicted soft-segmentation map ${\mm P}^{\trg{\mm x}}$, the entropy loss is:
    \begin{equation}
        \mc{L}_{ent}(\trg{\mm x}) = \frac{-1}{\log(C_s)}\sum_{h=1}^H\sum_{w=1}^W\sum_{c\in \mathcal{C}_s} \mm P^{\trg{\mm x}}[h,w,c] \log \mm P^{\trg{\mm x}}[h,w,c].
    \end{equation}

Differently, AdvEnt approaches global distribution alignment via adversarial training on the weighted self-information space.
We refer readers to the original work in~\cite{vu2019advent} for more details.

\begin{figure*}[t]
\captionsetup[subfigure]{labelformat=empty}
		\begin{subfigure}[t]{0.48\textwidth}
		    \centering
			\caption{(a) UDA problem with private classes}\vspace{-0.2cm}
		\includegraphics[clip=true,width=0.95\textwidth]{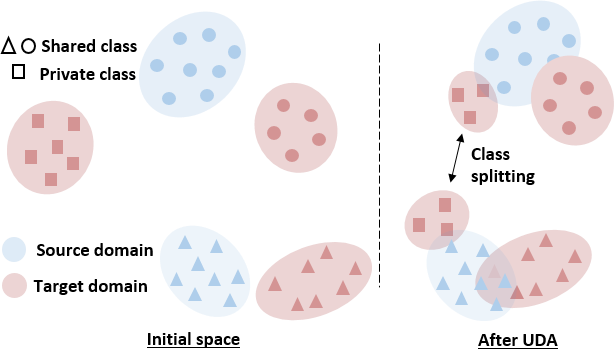}
		\end{subfigure}
		\begin{subfigure}[t]{0.48\textwidth}
		\centering
			\caption{(b) Zero-shot problem with domain gap}\vspace{-0.2cm}	\includegraphics[clip=true,width=0.77\textwidth]{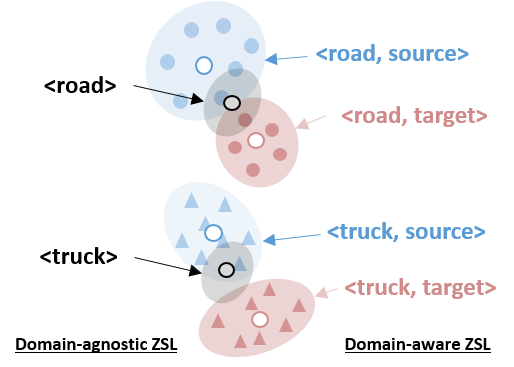}
		\end{subfigure}
	\caption{\textbf{Challenges of boundless UDA.} 
	(a) Existing UDA techniques, ignoring the presence of target private classes, tend to shift target-domain features to shared-class clusters. (b) Because of the domain gap, learning domain-agnostic mappings (black arrows) from word embedding space to visual space is sub-optimal as only mean modes (black circles) are captured. Instead, our domain-aware generative model can map to domain-specific modes (blue or red circles). }
	\label{fig:buda_challenges} 
\end{figure*}

As mentioned earlier, the mismatch between source and target label sets prevents the direct alignment between the visual distributions in source and target domains.
Figure~\ref{fig:buda_challenges}-(a) illustrates the problem caused by brute-force adoption of existing UDA techniques in the presence of private classes.
Indeed by design, AdvEnt, using global output-space alignment, and MinEnt, using global entropy aggregation, do not differentiate between shared and private classes, which eventually results in undesirable ``clusters'' of mixed shared-class and private-class features.

We propose a simple yet effective strategy based on MinEnt to mitigate the above concern.
	A segmentation network $F^{\,\textit{pre}}$ is first pre-trained only on the shared classes in the source domain.
	We then use $F^{\,\textit{pre}}$ to produce shared-class predictions on the target-domain training images, which may include both shared-class and private-class objects.
	We argue that, as $F^{\,\textit{pre}}$ has never observed private classes in the pre-training phase, top-confident predictions coming from~$F^{\,\textit{pre}}$ should mostly constitute of shared-class pixels.
	Effectively, applying entropy minimization solely on such top-confident pixels minimizes the risk of misalignment in the presence of private-class target-domain features.	
	We then fine-tune $F$ initialized with $F^{\,\textit{pre}}$, using the optimization objective:
	\begin{equation}
	\min_{\theta_F} \frac{1}{|\src{\cD}|}\sum_{\src{\mm x}\in\src{\cD}} \mathcal{L}_{\textit{seg}}(\src{\mm x}, \src{\mm y})  +  \frac{\lambda_{\textit{ent}}}{|\trg{\cD}|} \sum_{\trg{\mm x}\in\trg{\cD}}\hat{\mathcal{L}}_{\textit{ent}}(\trg{\mm x}),
	 \end{equation}
	 where $\theta_F$ are the parameters of $F$ and $\hat{\mathcal{L}}_{\textit{ent}}$ is derived from $\mathcal{L}_{\textit{ent}}$ by restricting the sum of the predictive entropies to the top-$k\%$ most-confident pixels according to  $F^{\,\textit{pre}}$. The parameter $\lambda_{\textit{ent}}$ controls the weight of this entropy term.
	Figure~\ref{fig:gc_gen} illustrates Step~$1$ operating on both source and target domains.

\subsection{BudaNet's Step~$2$: Domain-aware ZSL for private classes}
\label{sec:budanet_private}
The domain shift affects all classes, both shared and private ones.
Although for BUDA we assume the absence of private classes in the source domain, if there still appeared a private-class instance in a source-domain scene, its appearance should be notably different from target-domain instances.
For example, similar to `car' images, `tuk-tuk' images if taken in Paris should look different from those taken in India, due to distinct weather, illumination and overall appearance of each city.
We thus find it crucial to take domain information into account in Step~$2$ where the goal is to synthesize target-domain private-class features.
One may argue that after Step~$1$, the two domains are already aligned and the domain-invariant feature extractor $F_{\textit{feat}}$ should be ready as is for feature synthesis.
However, no existing UDA technique guarantees complete alignment across domains, which we thus do not expect either in our approach. 

We propose a novel domain-aware generative model, denoted $G_{\textit{DA}}$, which is trained to synthesize visual features for both source and target domains.
This generative model is modified such that it is conditioned not only on the class embedding but also on the domain (source or target).
Now, the modified model can output shared-class features for both source and target domains.
We note that, although our objective is to generate domain-aware features, those stemming from the same classes should be rather close.
Intuitively, we want our generator to mimic well the feature generator $F_{\textit{feat}}$, which produces close source-domain and target-domain feature distributions (as a result of the adaptation Step~$1$) yet still exhibiting certain discrepancies (as discussed above).

To this end, we leverage adversarial training.
In details, we introduce an additional discriminator $D$ trying to distinguish source-target generated features.
At train time, $D$ minimizes the binary cross-entropy classification loss; 
Given $\fakefeat^s$ and $\fakefeat^t$, two generated features for source (label $1$) and target (label $0$) domains, this loss reads: 
    \begin{equation}
    \mc{L}_{D}(\fakefeat^s, \fakefeat^t) = \mc{L}_{\textit{BCE}}(\fakefeat^s, 1) + \mc{L}_{\textit{BCE}}(\fakefeat^t, 0),
    \end{equation}
    with $\mc{L}_{\textit{BCE}}$ being the binary cross-entropy loss.
Meanwhile, the generative model trained with an additional adversarial loss $\mathcal{L}_{\textit{adv}}$ tries to confuse $D$.
Training is now based on the two losses: $\mathcal{L}_{\textit{GMMN}}$ as in Section~\ref{sec:based_pipeline} and the adversarial loss $\mathcal{L}_{\textit{adv}}$.
Given $\fakefeat^{t}$, a generated feature for target domain, this loss reads:
    \begin{equation}
    \mc{L}_{\textit{adv}}(\fakefeat^t) = \mc{L}_{\textit{BCE}}(\fakefeat^t, 1).
    \end{equation}
To train $G_{\textit{DA}}$, we use a weighting factor $\lambda_{\textit{adv}}$ for $\mc{L}_{\textit{adv}}$.

Generative training is supervised by real shared-class features coming from both source and target domains.
While in the source domain we can easily assign class labels to shared-class features using the ground-truth maps, in the target domain we must opt for a heuristic pseudo-labeling strategy.
Specifically, we run $F$ (pre-trained in Step $1$) on the unlabeled target-domain training set.
For each pixel, we assign the class with the highest prediction probability as its pseudo-label.
However, this labelling is error prone: it is not guaranteed to be correct on shared-class pixels, and cannot be correct on private-class pixels whose true labels are unknown to it. 
To mitigate such negative effects, only the top $p\%$ of the most confident pseudo-labels are kept; the rest are ignored during training.
Doing that, we improve pseudo-label quality for shared classes and reduce the number of retained private-class pixels.
In Figure~\ref{fig:gc_gen}, the strategy is illustrated as Step~$2$.

In Figure~\ref{fig:buda_challenges}-(b), we illustrate different outcomes of our proposed model as opposed to one without domain awareness.
As $G_{\textit{DA}}$ can capture domain-specific modes in the visual-feature space, we expect better generalization to private classes in the target domain.

\subsection{BudaNet's Step~$3$: Self-training with private classes} 
\label{sec:budanet_cls}
\begin{figure*}[]
    \centering
	\includegraphics[clip=true,width=0.8\textwidth]{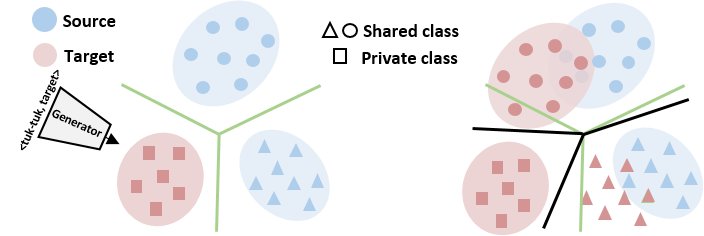}
	\caption{\textbf{Self-training with private classes}.
	(Left) Pre-training $F_{\textit{cls}}$ using aligned source-domain features of shared classes used in Step~$2$ and features generated for private classes, \textit{e.g.}, `tuk-tuk' indicated with squares. Solid lines illustrate the decision boundaries.
	(Right) Fine-tuning the whole segmenter $F$ using only pseudo-labeled target-class features -- blue points are there to contrast with the left figure but are not used to train $F$.}
	\label{fig:buda_step3} 
\end{figure*}

We start by only pre-training the final classification layer $F_{\textit{cls}}$ with the shared-class source-domain features used in Step $2$ and the generated private-class features coming from $G_{\textit{DA}}$ -- see Figure~\ref{fig:buda_step3} (left).
Once $F_{\textit{cls}}$ is trained, the network can now handle both shared and private objects at once, which effectively can be used to extract pseudo-labels on target-domain training set.
The whole segmentation network $F$ is then fine-tuned again with pseudo-labels -- similar to above, we only train on top confident ones.
Differently to the previous steps, the pseudo-labels now cover both shared and private classes.
Figures~\ref{fig:buda_step3} and \ref{fig:gc_gen}-(Step~$3$) illustrate this final self-training strategy.
In Figure~\ref{fig:buda_step3}-(right), the decision boundaries, learned after pre-training $F_{\textit{cls}}$ (blue lines) are shifted to new positions that are better adapted to the target domain after fine-tuning $F$ (black lines).

\section{Experiments}
In Section~\ref{sec:exp_detail}, we introduce our set-ups as well as some implementation details.
Section~\ref{sec:results} presents the main results, followed by ablation studies in Section~\ref{sec:ablation}.

\subsection{Experimental details}
\label{sec:exp_detail}

\vspace{-0.2cm}
\paragraph{BUDA scenarios}
To evaluate our approach, we define three BUDA scenarios: \textit{synthetic-2-real}, \textit{country-2-country} and \textit{dataset-2-dataset}. These very different set-ups allow investigating boundless adaptation from multiple angles of practical interest.

In the \textit{synthetic-2-real} set-up, available at train time are labelled synthetic data and unlabeled real images.
The zero-cost (and zero-risk) source-domain ground-truth acquisition makes this configuration especially appealing for semantic segmentation task.
We use the SYNTHIA-RAND-CITYSCAPES split of the SYNTHIA dataset~\cite{ros2016synthia} containing 9,400 synthesized images as for source domain.
The target-domain images are from the Cityscapes dataset~\cite{cordts2016cityscapes} with 2,975 images for training and 500 images for test.
In this set-up, we consider $16$ shared classes, previously used in some closed-set UDA works~\cite{hoffman2018cycada,tsai2018learning,vu2019advent}.
Differently, our target domain holds a private label-set of $3$ classes: `terrain', `truck' and `train'.

The \textit{country-2-country} set-up addresses the very practical use-case where a model trained using data from one region is deployed in other parts of the world.
In details, the source domain is defined by the Cityscapes data acquired in $50$ German cities while, for the target domain, we use images from the India driving dataset (IDD) \cite{varma2019idd}.
The geographic gap translates into a large visual gap on both the vehicles and the overall scenes' layout. 
While common vehicles like cars, trucks or buses exist almost everywhere in the two countries, only in India we observe auto-rickshaws.
It is also worth mentioning that class distributions are very different, \textit{i.e.} in Indian streets, the most frequent vehicles are motorbikes while in Germany, cars are more prevalent.
Such unique signatures of different countries can be challenging for applications like self-driving cars.
More into details, we use for training 2,975 real Cityscapes images with annotations along with 6,993 unlabeled IDD images.
There are~$19$ shared classes between the two datasets and~$2$ private classes only in IDD: `auto-rickshaw' (a.k.a. `tuk-tuk') and `animal'.
We evaluate models on $981$ IDD images for the total $21$ classes.

Our last BUDA set-up is \textit{dataset-2-dataset}: Pascal-VOC~\cite{everingham2015pascal} and MS-COCO~\cite{lin2014microsoft} as source and target domains.
As both were collected from the Internet, one may expect that the two datasets are very similar.
The big difference lies in the scene complexity and the level of annotation done in each: while Pascal-VOC images mostly have a single centered object with a limited $20$-class label-set, MS-COCO images are more complex with multiple objects exhaustively annotated.
Still, as the visual gap is small between the two datasets, we do not expect very significant performance drop in the shared classes.
The challenge clearly appears once we consider a target's private label set.
In detail, there are $1,464$ VOC images and ~$118$k COCO images used in training.
We consider, in the main experiment, $20$ shared classes and $5$ target private classes: `truck', `bench', `zebra', `giraffe' and `laptop'.
For validation, we use $5000$ COCO images which contain at least one of the $25$ classes. For the ablation study in Section~\ref{sec:ablation}, we also increased in this set-up the number of private classes from 5 to 20, something the two considered datasets permit while respecting the BUDA assumption that private classes do not appear in the source domain.

\vspace{-0.2cm}
\paragraph{Model implementation}
We adopt the DeepLabv3+ framework~\cite{chen2018encoder} built upon a ResNet-101~\cite{he2016deep} backbone. 
The SGD~\cite{bottou2010large} optimizer is used with polynomial learning rate decay with the base learning rate of $10\mathrm{e}-2$, weight decay $1\mathrm{e}{-4}$ and momentum $0.9$.

Generative models $G$ and $G_{\textit{DA}}$ are multi-layer perceptrons having a single hidden layer with leaky-ReLU non-linearity~\cite{maas2013rectifier} and dropout~\cite{srivastava2014dropout}.
We fix the number of hidden neurons as $256$.
Both generative models take as input a semantic classes' embedding of dimension $d_a = 300$ and a Gaussian noise vector of dimension $d_z=300$.
$G_{\textit{DA}}$ additionally accepts a $1-$dim vector specifying the domain (source or target).
Regarding the generative loss $\mathcal{L}_{\textit{GMMN}}$, we chose kernel bandwidths in $\{2, 5, 10, 20, 40, 60\}$.

For adversarial training of $G_{\textit{DA}}$, we introduce an additional binary classification discriminator $D$: a fully connected layer takes $256-$dim feature vectors and predicts the corresponding domain.
All generative models are trained using Adam optimizer~\cite{kingma2014adam} with a learning rate of $2\mathrm{e}{-4}$.

\paragraph{Evaluation metrics} We want to assess performance of both shared and private classes at test time.
Therefore, we adopt the GZSL evaluation protocol for semantic segmentation used in~\cite{bucher2019zero}.
The protocol considers three traditional base metrics in segmentation: pixel accuracy (PA), mean accuracy (MA) and mean intersection-over-union (mIoU), separately computed for shared and private sets.
Also reported are the harmonic means (hPA, hMA and hIoU) of shared and private results.
The harmonic means are used because the common metrics suffer from the performance bias of shared classes, while our aim is to achieve high accuracy for both shared and private classes~\cite{xian2018zero}.

\subsection{Results}
\label{sec:results}

Tables~\ref{tbl:cityscapes_res},~\ref{tbl:idd_res} and~\ref{tbl:coco_res} report our results in different setups.

In each experiment, we report performance for shared, private, and all classes.
``Supervised'' stands for the model trained with full supervision on the target domain.
``Oracle'' corresponds to the zero-shot framework trained on target-domain data using only shared labels; ``ZS3Net (only source)" is a similar network but only trained on source-domain data (based on ZS3Net in \cite{bucher2019zero}).
One straight-forward baseline is to directly apply a vanilla UDA technique, \textit{i.e.} MinEnt~\cite{vu2019advent}, in Step 1 without considering the private classes.
We denote ``ZS3Net\,+\,UDA'' this baseline.
We note that in the new BUDA setting, existing UDA techniques compare differently than in the classic setting, as later reported in Section~\ref{sec:ablation}; Indeed ``MinEnt'' performs better than a state-of-the-art adversarial training approach.
In addition, we consider the proposed strategy introduced in Section~\ref{sec:budanet_shared} as a stronger baseline, denoted ``ZS3Net\,+\,Adaptation'' in the result tables.

\paragraph{Synthetic-to-Real}

\begin{table*}[h!]
\centering
\caption{\textbf{Semantic segmentation performance on SYNTHIA$\rightarrow$Cityscapes.} Pixel accuracy (PA), mean accuracy (MA) and mean intersection-over-union (mIoU) on shared and private class sets, and their harmonic averages over the two class sets (hPA,hMA,hIoU). See text for method's discussion.}
\begin{tabular}{llrrrrrrrrrr}
\toprule
 & \multicolumn{3}{c}{Shared} && \multicolumn{3}{c}{Private} && \multicolumn{3}{c}{Overall}     \\
 \cline{2-4} \cline{6-8} \cline{10-12}\noalign{\smallskip}
Method               & PA    &MA    & mIoU  && PA    & MA    & mIoU    && hPA    & hMA   &  hIoU \\ \midrule
  Supervised & 95.1 & 68.1  & 61.7 && 58.1 & 60.0 & 48.2 && 72.1 & 63.8 &   54.1\\ 
  Oracle & 94.3 & 65.7  & 56.4 && 50.1 & 45.9 & 31.5 && 65.4 & 54.0  &  40.4 \\ \midrule
  ZS3Net (source only) \cite{bucher2019zero}  & 80.3 & 43.1  & 28.1 && 9.0 & 40.7 & 6.9  && 16.2  & 41.9 &  11.1  \\
   ZS3Net\,+\,UDA  & 85.3  &  42.4  & 30.1  && 12.5 & 46.8 & 8.2 &&  21.8 &  44.9 & 12.8  \\ 
  ZS3Net\,+\,Adaptation  & 89.9  &  42.6  & 35.0  && 18.6 & 51.1 & 8.9 &&  30.8 & 46.5  & 14.2  \\ 
 \rowcolor[gray]{.92}BudaNet & \textBF{93.0}  &  \textBF{46.0}  & \textBF{36.2}  && \textBF{26.9} & \textBF{58.7} & \textBF{17.0} && \textBF{41.7}  & \textBF{51.6}   & \textBF{23.1} \\
\bottomrule
\end{tabular}
\label{tbl:cityscapes_res}
\end{table*}

\begin{figure*}[h!]
    \captionsetup[subfigure]{labelformat=empty}
	\begin{center}
     \begin{subfigure}[t]{0.24\textwidth}\centering 
						\caption{Input image}\vspace{-0.2cm}
			\includegraphics[trim=0cm 0.0cm 0cm 0.0cm, clip=true,width=.98\textwidth]{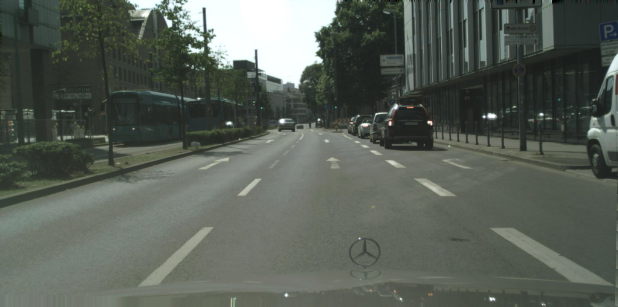}
		\end{subfigure} 
		\begin{subfigure}[t]{0.24\textwidth}\centering
			\caption{GT}\vspace{-0.2cm}
			\includegraphics[trim=0cm 0.0cm 0cm 0.0cm, clip=true,width=.98\textwidth]{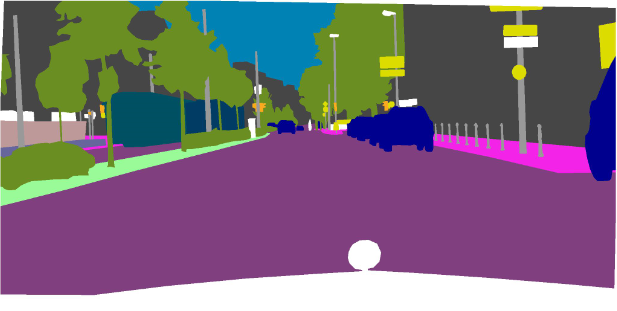}
		\end{subfigure} 
		\begin{subfigure}[t]{0.24\textwidth}\centering
			\caption{ZS3Net}\vspace{-0.2cm}
			\includegraphics[trim=0cm 0.0cm 0cm 0.0cm, clip=true,width=.98\textwidth]{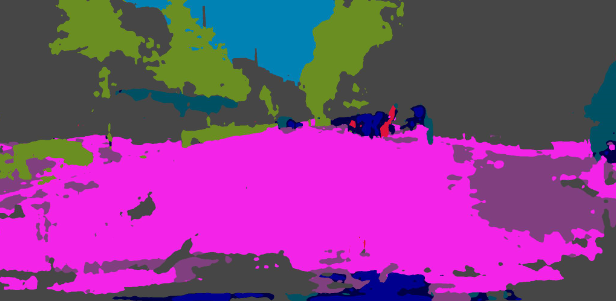}
		\end{subfigure} 
		\begin{subfigure}[t]{0.24\textwidth}\centering
		\caption{BudaNet}\vspace{-0.2cm}
			\includegraphics[trim=0cm 0.0cm 0cm 0.0cm, clip=true,width=.98\textwidth]{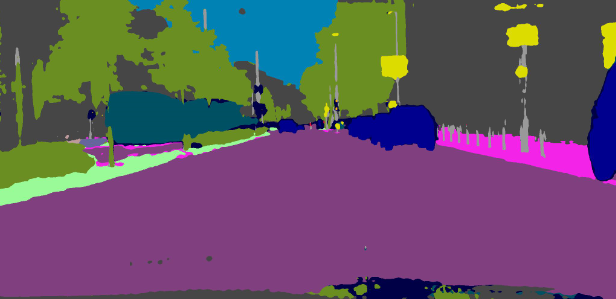}
		\end{subfigure}\\ 
 		\begin{subfigure}[t]{0.24\textwidth}\centering 
 			\includegraphics[trim=0cm 0.0cm 0cm 0.0cm, clip=true,width=.98\textwidth]{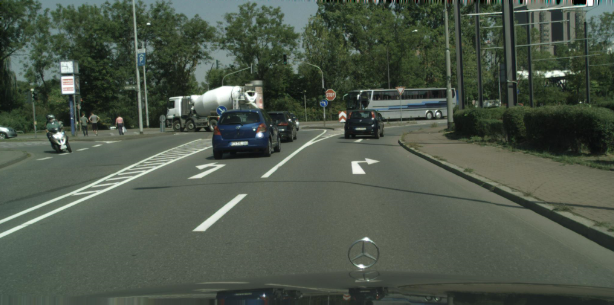}
		\end{subfigure} 
 		\begin{subfigure}[t]{0.24\textwidth}\centering
 		\includegraphics[trim=0cm 0.0cm 0cm 0.0cm, clip=true,width=.98\textwidth]{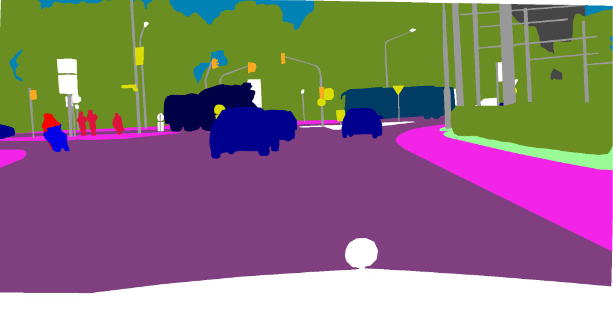}
		\end{subfigure} 
 		\begin{subfigure}[t]{0.24\textwidth}\centering
 		\includegraphics[trim=0cm 0.0cm 0cm 0.0cm, clip=true,width=.98\textwidth]{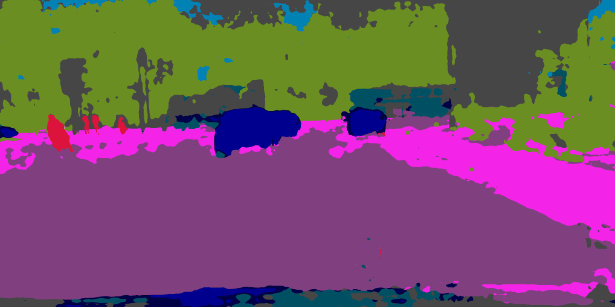}
		\end{subfigure} 
 		\begin{subfigure}[t]{0.24\textwidth}\centering
 		\includegraphics[trim=0cm 0.0cm 0cm 0.0cm, clip=true,width=.98\textwidth]{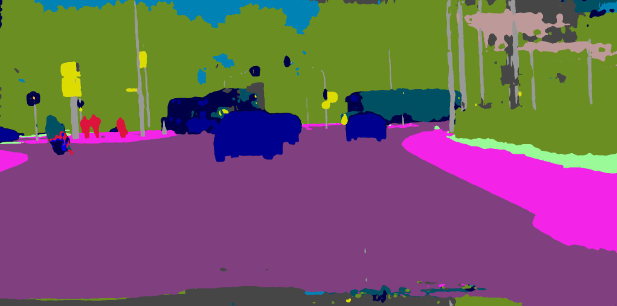}
		\end{subfigure}\\
		\begin{subfigure}[t]{0.24\textwidth}\centering 
 			\includegraphics[trim=0cm 0.0cm 0cm 0.0cm, clip=true,width=.98\textwidth]{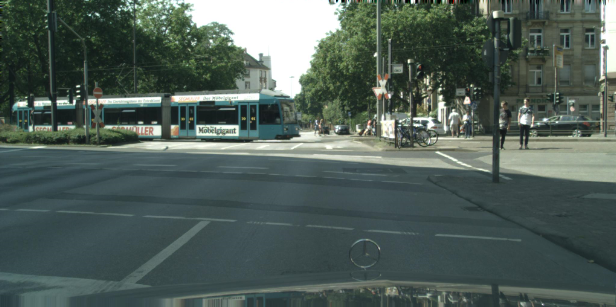}
		\end{subfigure} 
 		\begin{subfigure}[t]{0.24\textwidth}\centering
 			\includegraphics[trim=0cm 0.0cm 0cm 0.0cm, clip=true,width=.98\textwidth]{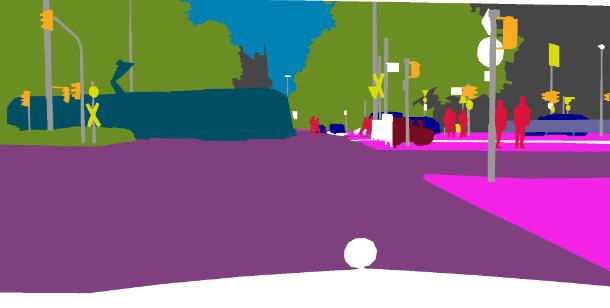}
		\end{subfigure} 
 		\begin{subfigure}[t]{0.24\textwidth}\centering
 		\includegraphics[trim=0cm 0.0cm 0cm 0.0cm, clip=true,width=.98\textwidth]{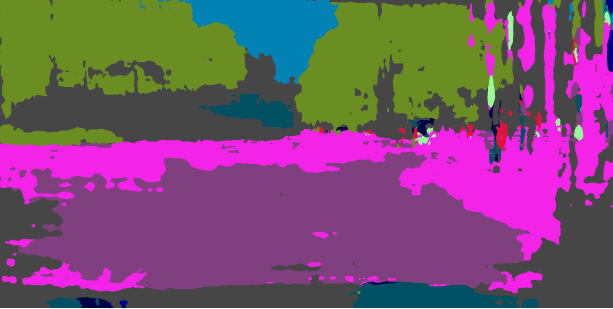}
		\end{subfigure} 
 		\begin{subfigure}[t]{0.24\textwidth}\centering
 			\includegraphics[trim=0cm 0.0cm 0cm 0.0cm, clip=true,width=.98\textwidth]{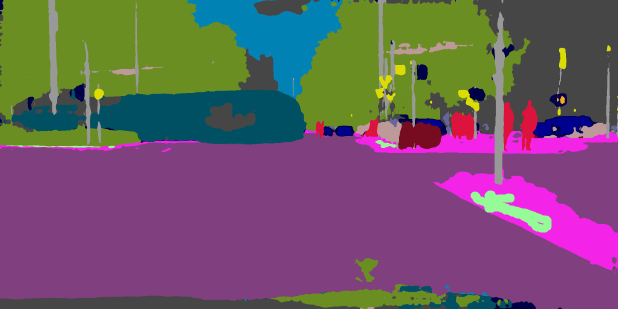}
		\end{subfigure}\\
		
		\begin{subfigure}[t]{0.24\textwidth}\centering 
 		\includegraphics[trim=0cm 0.0cm 0cm 0.0cm, clip=true,width=.98\textwidth]{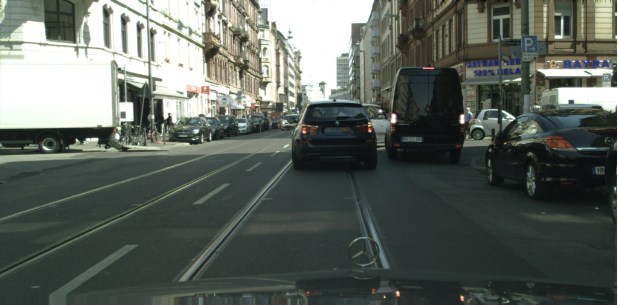}
		\end{subfigure} 
 		\begin{subfigure}[t]{0.24\textwidth}\centering
 			\includegraphics[trim=0cm 0.0cm 0cm 0.0cm, clip=true,width=.98\textwidth]{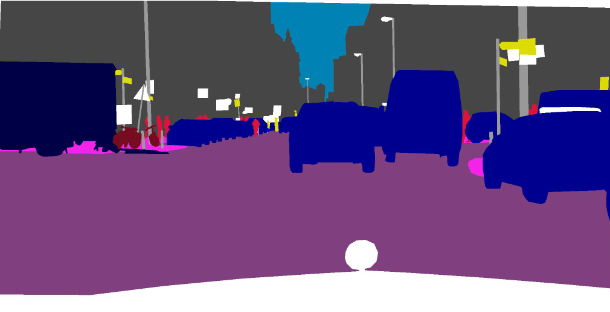}
		\end{subfigure} 
 		\begin{subfigure}[t]{0.24\textwidth}\centering
 		\includegraphics[trim=0cm 0.0cm 0cm 0.0cm, clip=true,width=.98\textwidth]{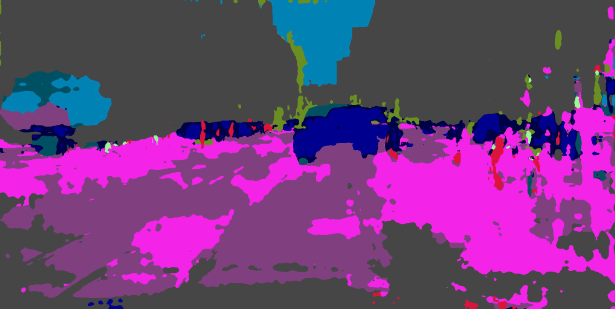}
		\end{subfigure} 
 		\begin{subfigure}[t]{0.24\textwidth}\centering
 			\includegraphics[trim=0cm 0.0cm 0cm 0.0cm, clip=true,width=.98\textwidth]{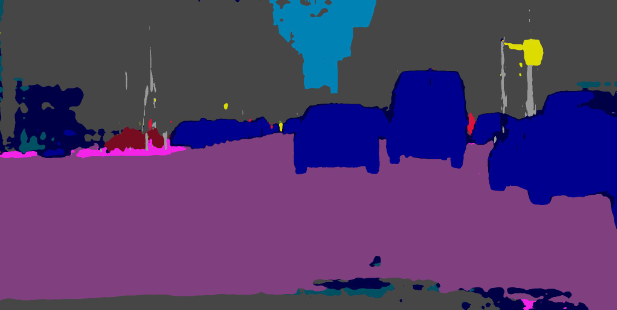}
		\end{subfigure}\\
		
		\begin{subfigure}[t]{0.24\textwidth}\centering 
 		\includegraphics[trim=0cm 0.0cm 0cm 0.0cm, clip=true,width=.98\textwidth]{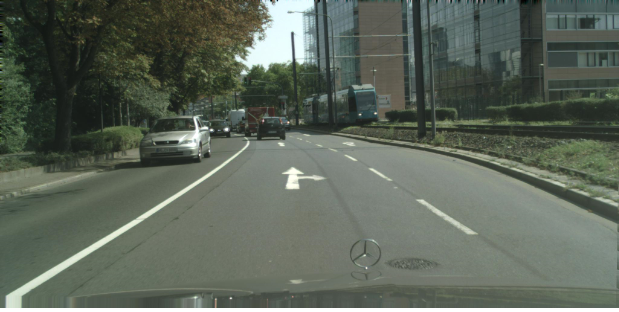}
		\end{subfigure} 
 		\begin{subfigure}[t]{0.24\textwidth}\centering
 			\includegraphics[trim=0cm 0.0cm 0cm 0.0cm, clip=true,width=.98\textwidth]{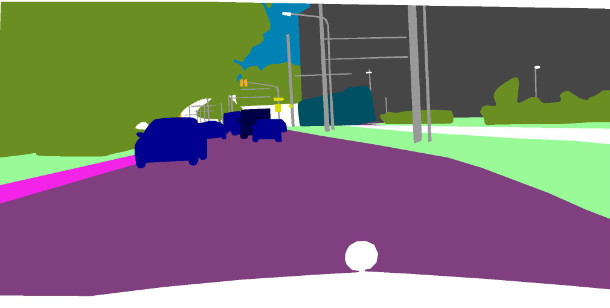}
		\end{subfigure} 
 		\begin{subfigure}[t]{0.24\textwidth}\centering
 		\includegraphics[trim=0cm 0.0cm 0cm 0.0cm, clip=true,width=.98\textwidth]{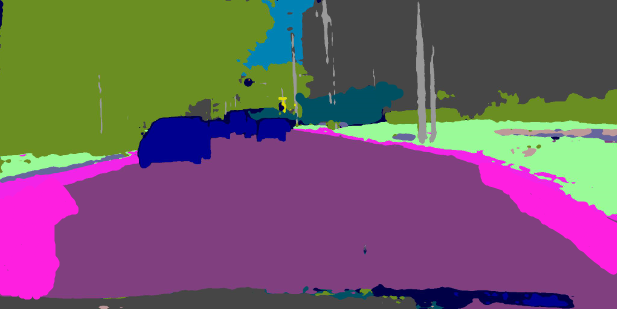}
		\end{subfigure} 
 		\begin{subfigure}[t]{0.24\textwidth}\centering
 			\includegraphics[trim=0cm 0.0cm 0cm 0.0cm, clip=true,width=.98\textwidth]{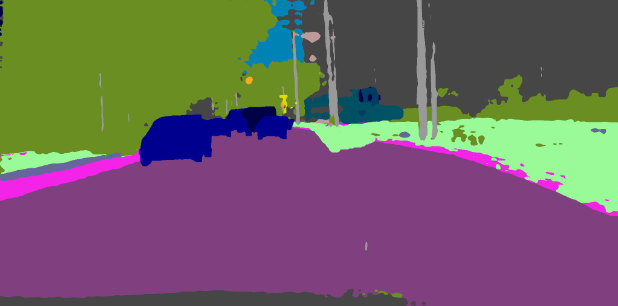}
		\end{subfigure}\\
		
		\begin{subfigure}[t]{0.24\textwidth}\centering 
 		\includegraphics[trim=0cm 0.0cm 0cm 0.0cm, clip=true,width=.98\textwidth]{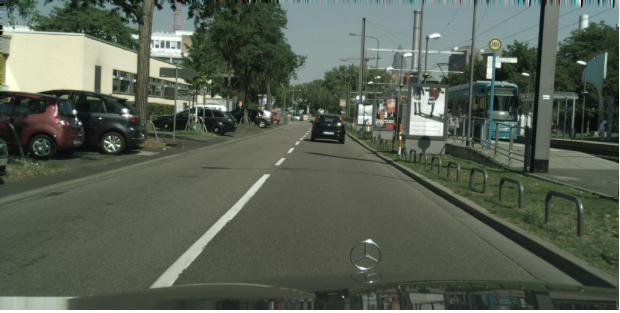}
		\end{subfigure} 
 		\begin{subfigure}[t]{0.24\textwidth}\centering
 			\includegraphics[trim=0cm 0.0cm 0cm 0.0cm, clip=true,width=.98\textwidth]{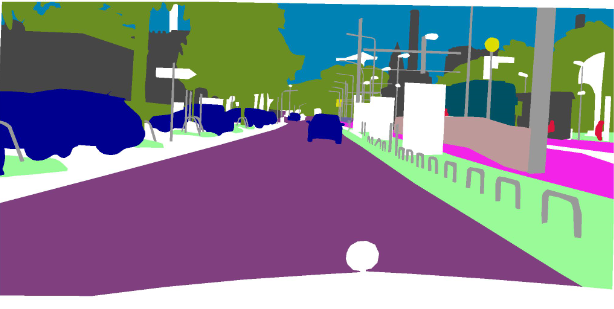}
		\end{subfigure} 
 		\begin{subfigure}[t]{0.24\textwidth}\centering
 		\includegraphics[trim=0cm 0.0cm 0cm 0.0cm, clip=true,width=.98\textwidth]{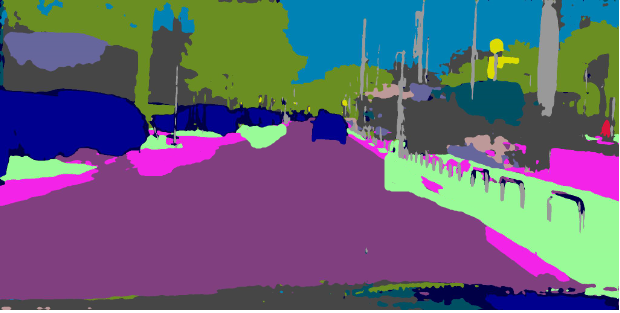}
		\end{subfigure} 
 		\begin{subfigure}[t]{0.24\textwidth}\centering
 		\includegraphics[trim=0cm 0.0cm 0cm 0.0cm, clip=true,width=.98\textwidth]{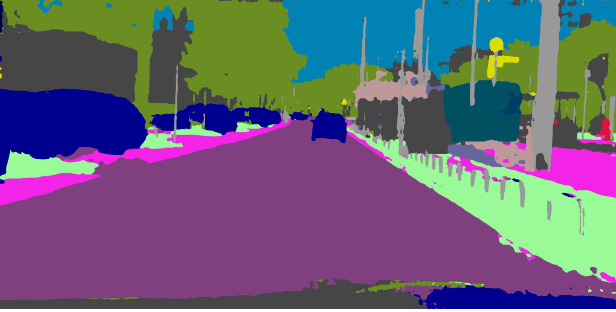}
		\end{subfigure}\\
	\end{center}
	\caption{\textbf{Qualitative results on Cityscapes}. 
	 The first and second columns show input images and corresponding segmentation ground truth. The third and fourth columns visualize results produced by ZS3Net and BudaNet.
	Private target classes: \setlength{\fboxsep}{1pt}\colorbox{col_terrain}{\textcolor{white}{terrain}}, \setlength{\fboxsep}{1pt}\colorbox{col_truck}{\textcolor{white}{truck}},
	\setlength{\fboxsep}{1pt}\colorbox{col_train}{\textcolor{white}{train}}. Some shared classes: \setlength{\fboxsep}{1pt}\colorbox{col_road}{\textcolor{white}{road}}, \setlength{\fboxsep}{1pt}\colorbox{col_sidewalk}{\textcolor{white}{side walk}}, \setlength{\fboxsep}{1pt}\colorbox{col_car}{\textcolor{white}{car}},
	\setlength{\fboxsep}{1pt}\colorbox{col_person}{\textcolor{white}{person}},
	\setlength{\fboxsep}{1pt}\colorbox{col_moto}{\textcolor{white}{motorbike}},
	\setlength{\fboxsep}{1pt}\colorbox{col_tree}{\textcolor{white}{tree}},
	\setlength{\fboxsep}{1pt}\colorbox{col_building}{\textcolor{white}{building}}.}
	\label{fig:qual_supmat_cityscapes}
\end{figure*}
\begin{table*}[h!]
\centering
\caption{\textbf{Semantic segmentation performance on Cityscapes$\rightarrow$IDD.} Pixel accuracy (PA), mean accuracy (MA) and mean intersection-over-union (mIoU) on shared and private class sets, and their harmonic averages over the two class sets (hPA,hMA,hIoU). See text for method's discussion.}
\begin{tabular}{llrrrrrrrrrr}
\toprule
 & \multicolumn{3}{c}{Shared} && \multicolumn{3}{c}{Private} && \multicolumn{3}{c}{Overall}     \\
 \cline{2-4} \cline{6-8} \cline{10-12}\noalign{\smallskip}
Method           & PA    &MA    & mIoU  && PA    & MA    & mIoU    && hPA    & hMA   &  hIoU \\ \midrule 
  Supervised & 92.8 & 65.1  & 56.9 && 57.3 & 61.9 & 48.0  && 70.9 & 63.5 &   52.1 \\
     Oracle & 91.7 & 63.0  & 53.1 && 47.2 & 41.8 & 28.8  && 62.3 & 50.3 &  37.3 \\ \midrule
  ZS3Net (source only) \cite{bucher2019zero}   & 81.0 & 43.8  & 29.2 && 9.3 & 41.0 & 7.9 && 16.7 & 42.4  &  12.4 \\
  ZS3Net\,+\,UDA  &  86.8 &  40.0  & 32.4  && 13.8 & 45.9 & 8.1 && 23.8  & 42.7  & 13.0  \\ 
  ZS3Net\,+\,Adaptation  & 88.3 & 40.5   & 32.7 && 15.9 & 47.0 & 8.6 && 26.9 & 43.5 &  13.6 \\ 
 \rowcolor[gray]{.92}BudaNet & \textBF{92.0}  &  \textBF{47.2}  & \textBF{37.3}  && \textBF{28.6} & \textBF{58.9} & \textBF{18.5} && \textBF{43.6}  & \textBF{52.4}   &  \textBF{24.7} \\
\bottomrule
\end{tabular}
\label{tbl:idd_res}
\end{table*}
\begin{figure*}[h!]
    \captionsetup[subfigure]{labelformat=empty}
	\begin{center}
 	    \begin{subfigure}[t]{0.24\textwidth}\centering 
 						\caption{Input image}\vspace{-0.2cm}
 			\includegraphics[trim=0cm 0.0cm 0cm 0.0cm, clip=true,width=.98\textwidth]{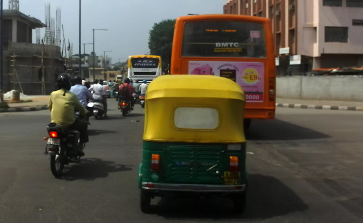}
 		\end{subfigure} 
 		\begin{subfigure}[t]{0.24\textwidth}\centering
 			\caption{GT}\vspace{-0.2cm}
 			\includegraphics[trim=0cm 0.1cm 0cm 0.0cm, clip=true,width=.98\textwidth]{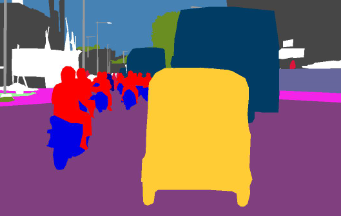}
 		\end{subfigure} 
 		\begin{subfigure}[t]{0.24\textwidth}\centering
 			\caption{ZS3Net}\vspace{-0.2cm}
 			\includegraphics[trim=0cm 0.2cm 0cm 0.0cm, clip=true,width=.98\textwidth]{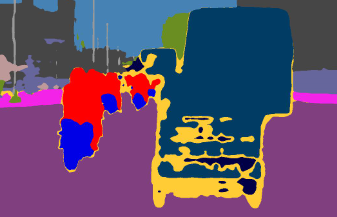}
 		\end{subfigure} 
 		\begin{subfigure}[t]{0.24\textwidth}\centering
 		\caption{BudaNet}\vspace{-0.2cm}
 			\includegraphics[trim=0cm 0.2cm 0cm 0.0cm, clip=true,width=.98\textwidth]{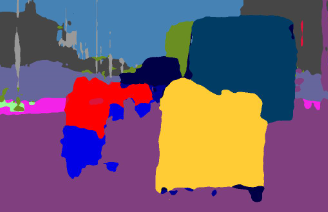}
 		\end{subfigure}\\ 
 			\begin{subfigure}[t]{0.24\textwidth}\centering 
 			\includegraphics[trim=0cm 0.0cm 0cm 0.0cm, clip=true,width=.98\textwidth]{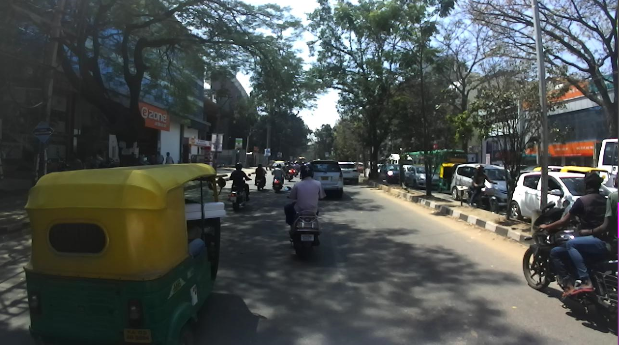}
		\end{subfigure} 
 		\begin{subfigure}[t]{0.24\textwidth}\centering
 			\includegraphics[trim=0cm 0.0cm 0cm 0.0cm, clip=true,width=.98\textwidth]{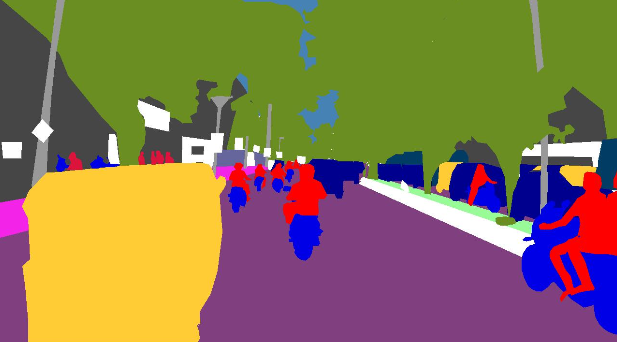}
		\end{subfigure} 
 		\begin{subfigure}[t]{0.24\textwidth}\centering
 		\includegraphics[trim=0cm 0.0cm 0cm 0.0cm, clip=true,width=.98\textwidth]{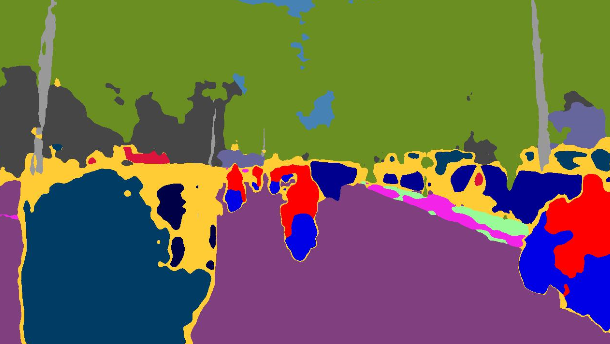}
		\end{subfigure} 
 		\begin{subfigure}[t]{0.24\textwidth}\centering
 			\includegraphics[trim=0cm 0.0cm 0cm 0.0cm, clip=true,width=.98\textwidth]{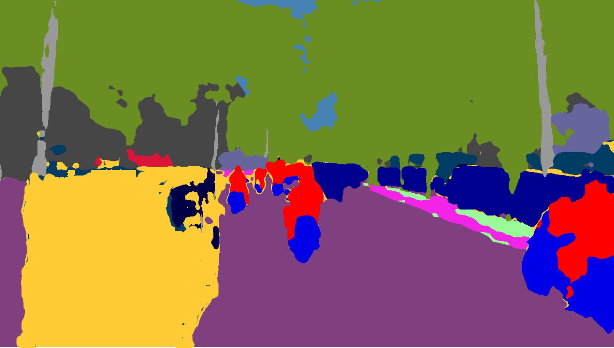}
		\end{subfigure}\\
		\begin{subfigure}[t]{0.24\textwidth}\centering 
			\includegraphics[trim=0cm 0.0cm 0cm 0.0cm, clip=true,width=.98\textwidth]{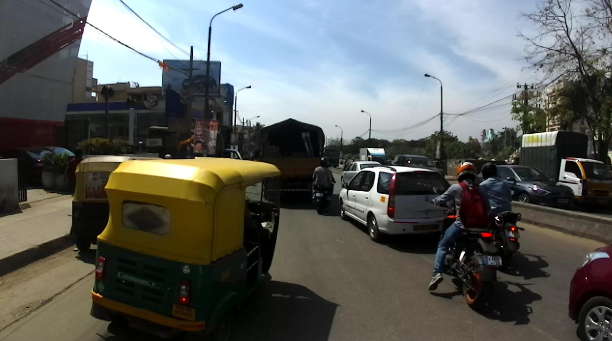}
		\end{subfigure} 
		\begin{subfigure}[t]{0.24\textwidth}\centering
			\includegraphics[trim=0cm 0.0cm 0cm 0.0cm, clip=true,width=.98\textwidth]{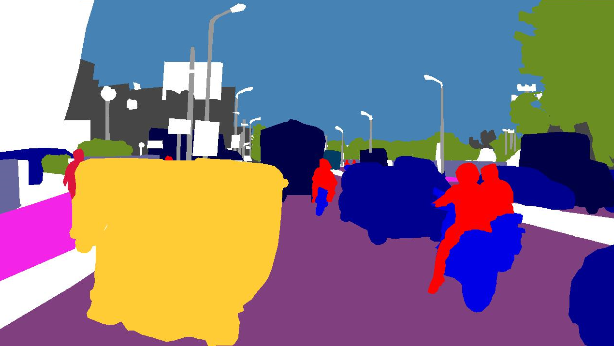}
		\end{subfigure} 
		\begin{subfigure}[t]{0.24\textwidth}\centering
			\includegraphics[trim=0cm 0.0cm 0cm 0.0cm, clip=true,width=.98\textwidth]{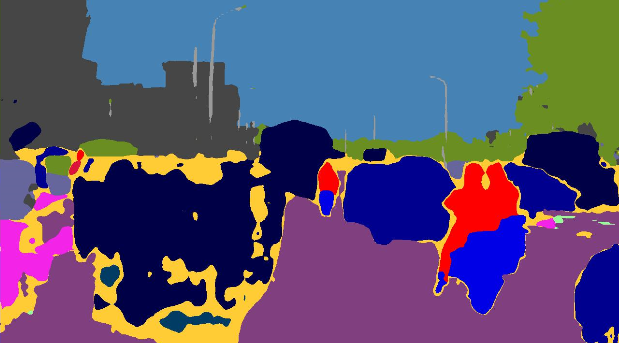}
		\end{subfigure} 
		\begin{subfigure}[t]{0.24\textwidth}\centering
			\includegraphics[trim=0cm 0.0cm 0cm 0.0cm, clip=true,width=.98\textwidth]{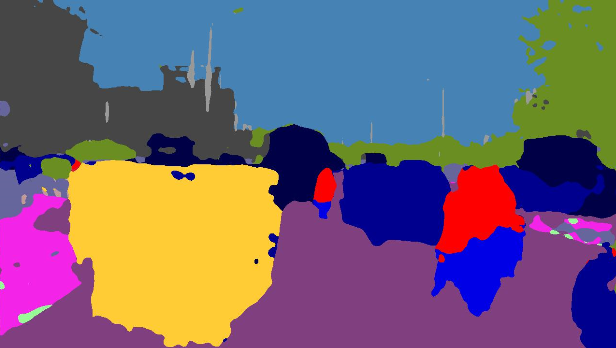}
		\end{subfigure}\\
			\begin{subfigure}[t]{0.24\textwidth}\centering 
 			\includegraphics[trim=0cm 0.0cm 0cm 0.0cm, clip=true,width=.98\textwidth]{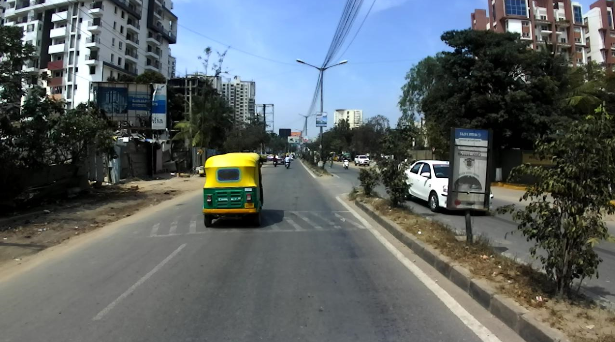}
		\end{subfigure} 
 		\begin{subfigure}[t]{0.24\textwidth}\centering
 			\includegraphics[trim=0cm 0.0cm 0cm 0.0cm, clip=true,width=.98\textwidth]{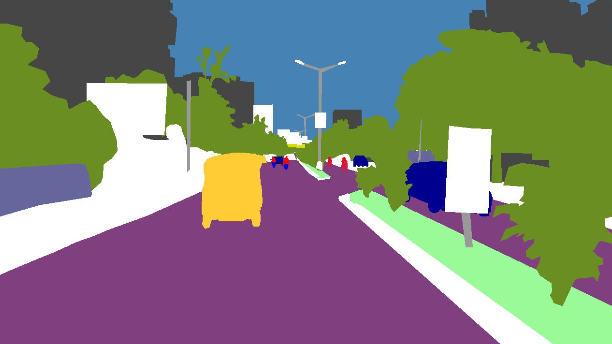}
		\end{subfigure} 
 		\begin{subfigure}[t]{0.24\textwidth}\centering
 		\includegraphics[trim=0cm 0.0cm 0cm 0.0cm, clip=true,width=.98\textwidth]{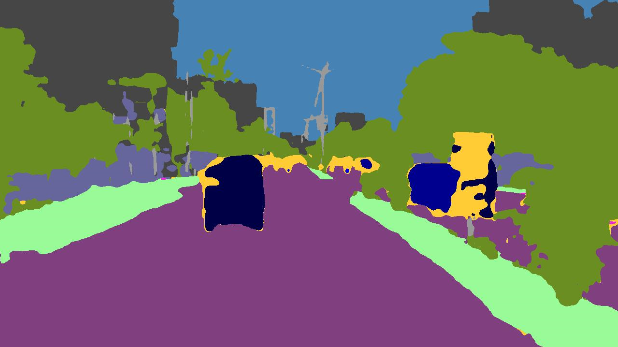}
		\end{subfigure} 
 		\begin{subfigure}[t]{0.24\textwidth}\centering
 			\includegraphics[trim=0cm 0.0cm 0cm 0.0cm, clip=true,width=.98\textwidth]{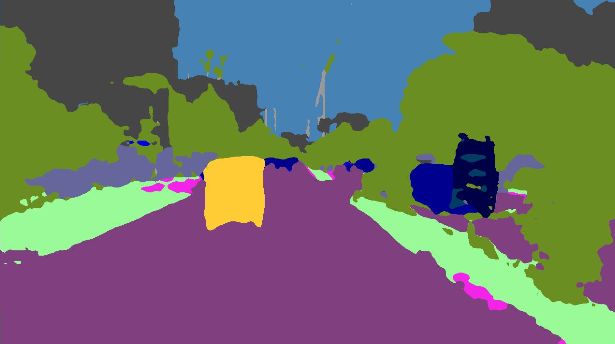}
		\end{subfigure}\\
		\begin{subfigure}[t]{0.24\textwidth}\centering 
 			\includegraphics[trim=0cm 0.0cm 0cm 0.0cm, clip=true,width=.98\textwidth]{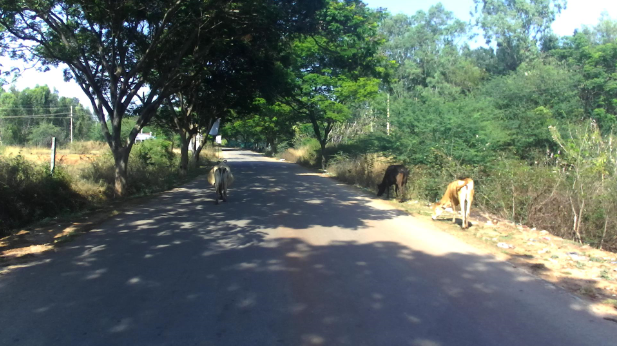}
		\end{subfigure} 
 		\begin{subfigure}[t]{0.24\textwidth}\centering
 			\includegraphics[trim=0cm 0.0cm 0cm 0.0cm, clip=true,width=.98\textwidth]{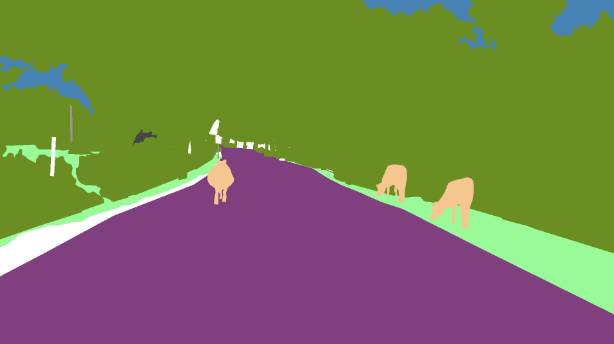}
		\end{subfigure} 
 		\begin{subfigure}[t]{0.24\textwidth}\centering
 		\includegraphics[trim=0cm 0.0cm 0cm 0.0cm, clip=true,width=.98\textwidth]{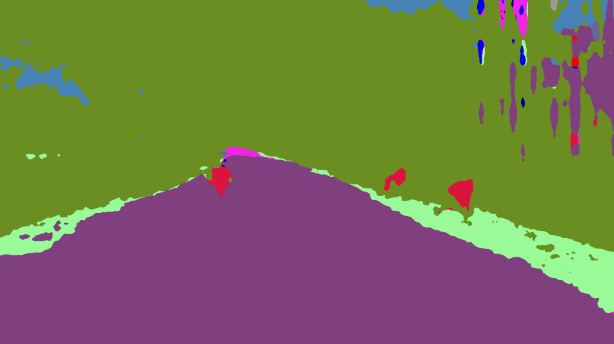}
		\end{subfigure} 
 		\begin{subfigure}[t]{0.24\textwidth}\centering
 			\includegraphics[trim=0cm 0.0cm 0cm 0.0cm, clip=true,width=.98\textwidth]{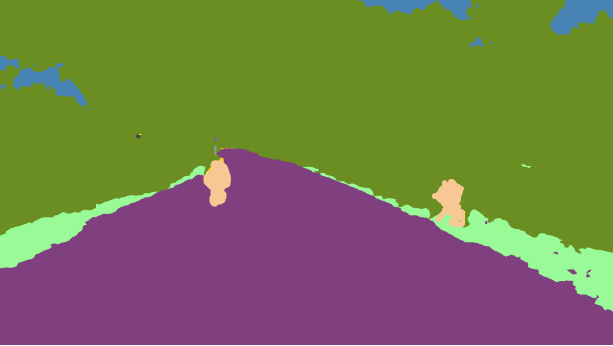}
		\end{subfigure}
	\end{center}
	\vspace{-0.2cm}
	\caption{\textbf{Qualitative results on IDD}. 
	The first and second columns show input images and corresponding segmentation ground truth. The third and fourth columns visualize results produced by ZS3Net and BudaNet.
	Private target classes:
	\setlength{\fboxsep}{1pt}\colorbox{col_tuk}{\textcolor{white}{tuk-tuk}},
	\setlength{\fboxsep}{1pt}\colorbox{col_animal}{\textcolor{white}{animal}}.
	Some shared classes:
	\setlength{\fboxsep}{1pt}\colorbox{col_truck}{\textcolor{white}{truck}}, \setlength{\fboxsep}{1pt}\colorbox{col_road}{\textcolor{white}{road}}, \setlength{\fboxsep}{1pt}\colorbox{col_sidewalk}{\textcolor{white}{side walk}}, \setlength{\fboxsep}{1pt}\colorbox{col_car}{\textcolor{white}{car}},
	\setlength{\fboxsep}{1pt}\colorbox{col_person}{\textcolor{white}{person}},
	\setlength{\fboxsep}{1pt}\colorbox{col_moto}{\textcolor{white}{motorbike}},
	\setlength{\fboxsep}{1pt}\colorbox{col_tree}{\textcolor{white}{tree}},
	\setlength{\fboxsep}{1pt}\colorbox{col_building}{\textcolor{white}{building}}.}
	\label{fig:qual_supmat_idd}
	\vspace{-0.2cm}
\end{figure*}

\begin{table*}[h!]
\centering
\caption{\textbf{Semantic segmentation performance on Pascal-VOC$\rightarrow$MS-COCO.} 
Pixel accuracy (PA), mean accuracy (MA) and mean intersection-over-union (mIoU) on shared and private class sets, and their harmonic averages over the two class sets (hPA,hMA,hIoU). See text for method's discussion.}
\begin{tabular}{llrrrrrrrrrr}
\toprule
 & \multicolumn{3}{c}{Shared} && \multicolumn{3}{c}{Private} && \multicolumn{3}{c}{Overall}     \\
 \cline{2-4} \cline{6-8} \cline{10-12}\noalign{\smallskip}
Method & PA    &MA    & mIoU  && PA    & MA    & mIoU    && hPA    & hMA   &  hIoU \\ \midrule
  Supervised & 94.8 & 68.8  & 70.1 && 70.5 & 61.0 & 64.3  && 80.9 & 64.7 &  67.1  \\ 
  Oracle & 93.8 & 68.6  & 68.0 && 61.6 & 57.7 & 39.9  && 74.4  & 62.8 &  50.3   \\ \midrule
  ZS3Net (source only) \cite{bucher2019zero}   & 92.0 & 67.3  & 63.3 && 29.9 & 51.4 & 17.3  && 45.1 & 58.3 & 27.2 \\
  ZS3Net\,+\,UDA  & 92.3  &  68.0  & 63.3  &&32.3 & 50.3 & 19.5 &&  52.0 &  47.9 &  29.8 \\ 
  ZS3Net\,+\,Adaptation & 92.3 & 68.2  & 63.8 && 36.2 & 54.1 & 20.9  && 52.0 & 60.3 & 31.5   \\ 
\rowcolor[gray]{.92}BudaNet & \textBF{93.1} & \textBF{68.4}  & \textBF{65.0} && \textBF{38.5} & \textBF{56.5} & \textBF{23.8}  && \textBF{54.5} & \textBF{61.9} &   \textBF{34.8}   \\
\bottomrule
\end{tabular}
\label{tbl:coco_res}
\end{table*}

\begin{figure*}[h!]
    \captionsetup[subfigure]{labelformat=empty}
	\begin{center}
		\begin{subfigure}[t]{0.24\textwidth}\centering 
		\caption{Input image}\vspace{-0.2cm}
 			\includegraphics[trim=0cm 0.0cm 0cm 0.0cm, clip=true,width=.98\textwidth]{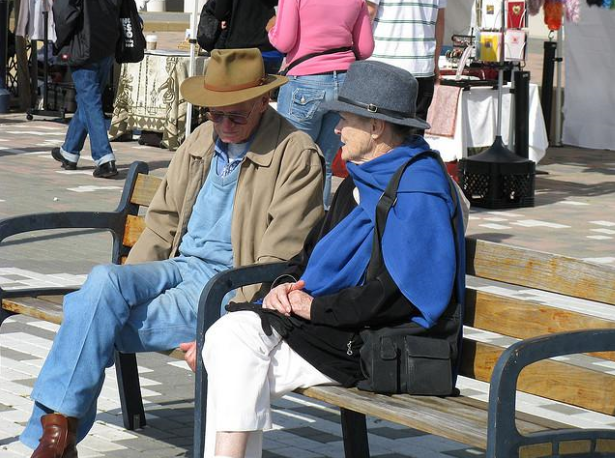}
		\end{subfigure} 
 		\begin{subfigure}[t]{0.24\textwidth}\centering
 		\caption{GT}\vspace{-0.2cm}
 			\includegraphics[trim=0cm 0.0cm 0cm 0.0cm, clip=true,width=.98\textwidth]{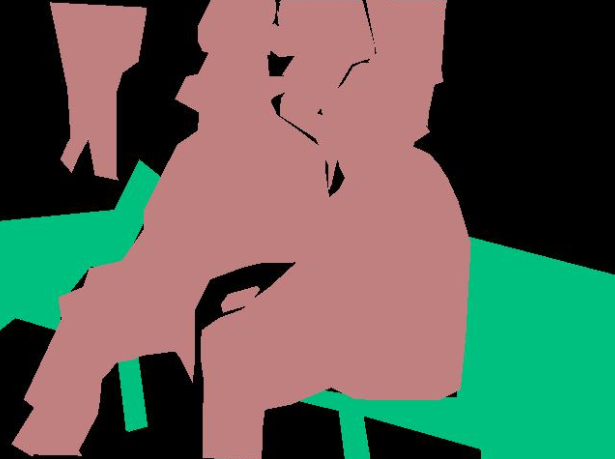}
		\end{subfigure} 
 		\begin{subfigure}[t]{0.24\textwidth}\centering
 			\caption{ZS3Net}\vspace{-0.2cm}
 		\includegraphics[trim=0cm 0.0cm 0cm 0.0cm, clip=true,width=.98\textwidth]{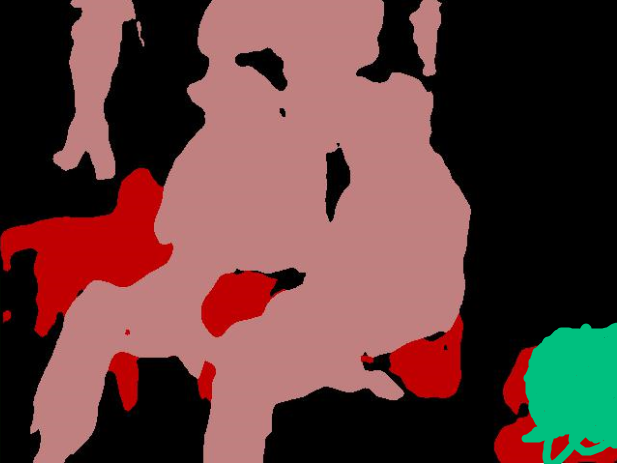}
		\end{subfigure} 
 		\begin{subfigure}[t]{0.24\textwidth}\centering
 		    \caption{BudaNet}\vspace{-0.2cm}
 			\includegraphics[trim=0cm 0.0cm 0cm 0.0cm, clip=true,width=.98\textwidth]{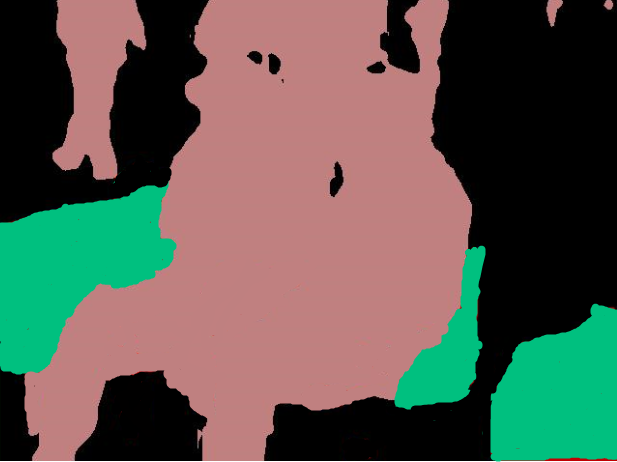}
		\end{subfigure}\\
			\begin{subfigure}[t]{0.24\textwidth}\centering 
			\includegraphics[trim=0cm 1.5cm 0cm 0.0cm, clip=true,width=.98\textwidth]{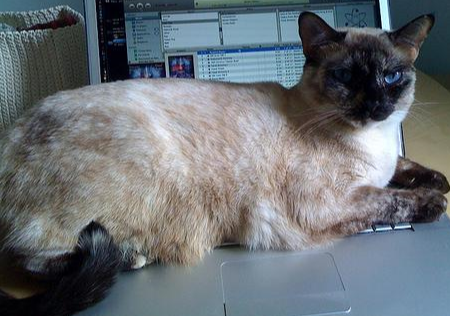}
		\end{subfigure} 
		\begin{subfigure}[t]{0.24\textwidth}\centering
			\includegraphics[trim=0cm 1.0cm 0cm 0.0cm, clip=true,width=.98\textwidth]{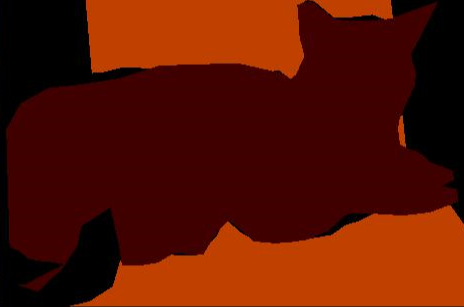}
		\end{subfigure} 
		\begin{subfigure}[t]{0.24\textwidth}\centering
			\includegraphics[trim=0cm 1.0cm 0cm 0.0cm, clip=true,width=.98\textwidth]{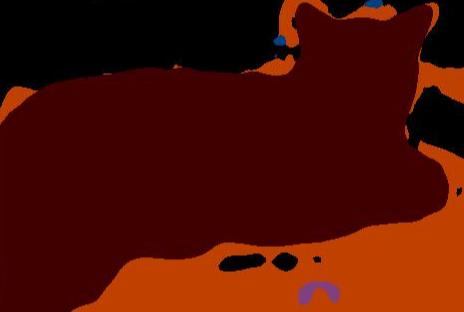}
		\end{subfigure} 
		\begin{subfigure}[t]{0.24\textwidth}\centering
			\includegraphics[trim=0cm 1.0cm 0cm 0.0cm, clip=true,width=.98\textwidth]{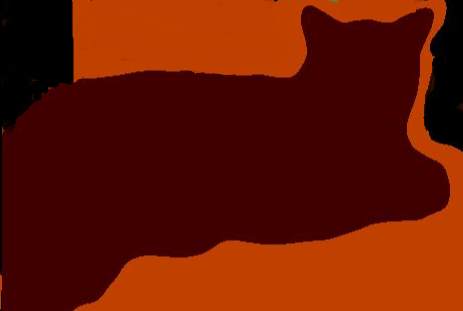}
		\end{subfigure}\\
	\begin{subfigure}[t]{0.24\textwidth}\centering 
 			\includegraphics[trim=0cm 0.0cm 0cm 0.0cm, clip=true,width=.98\textwidth]{./qual_coco/sup3_image.PNG}
		\end{subfigure} 
 		\begin{subfigure}[t]{0.24\textwidth}\centering
 			\includegraphics[trim=0cm 0.0cm 0cm 0.0cm, clip=true,width=.98\textwidth]{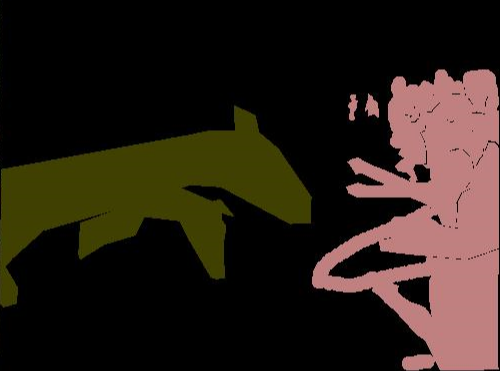}
		\end{subfigure} 
 		\begin{subfigure}[t]{0.24\textwidth}\centering
 		\includegraphics[trim=0cm 0.0cm 0cm 0.0cm, clip=true,width=.98\textwidth]{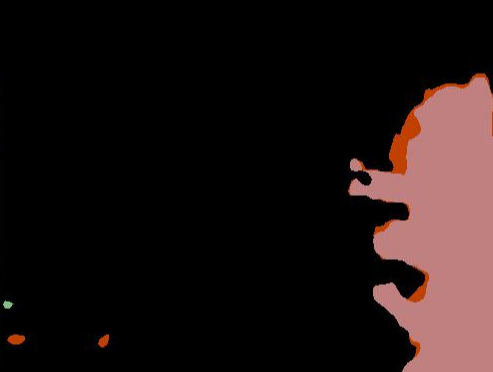}
		\end{subfigure} 
 		\begin{subfigure}[t]{0.24\textwidth}\centering
 			\includegraphics[trim=0cm 0.0cm 0cm 0.0cm, clip=true,width=.98\textwidth]{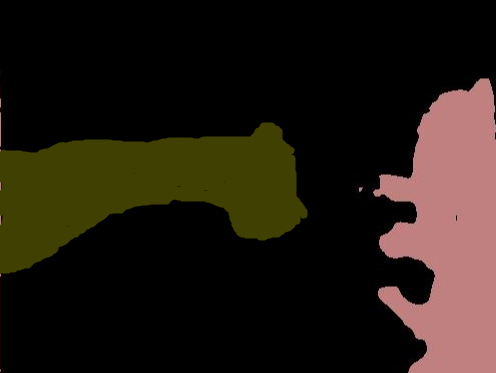}
		\end{subfigure}\\
			\begin{subfigure}[t]{0.24\textwidth}\centering 
 			\includegraphics[trim=0cm 0.0cm 0cm 0.0cm, clip=true,width=.98\textwidth]{./qual_coco/sup2_image.PNG}
		\end{subfigure} 
 		\begin{subfigure}[t]{0.24\textwidth}\centering
 			\includegraphics[trim=0cm 0.0cm 0cm 0.0cm, clip=true,width=.98\textwidth]{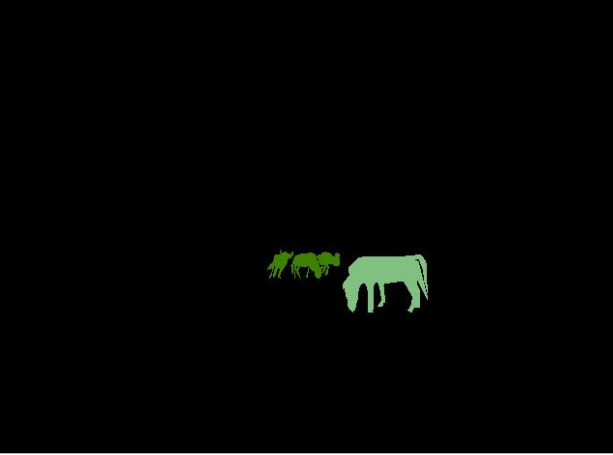}
		\end{subfigure} 
 		\begin{subfigure}[t]{0.24\textwidth}\centering
 		\includegraphics[trim=0cm 0.0cm 0cm 0.0cm, clip=true,width=.98\textwidth]{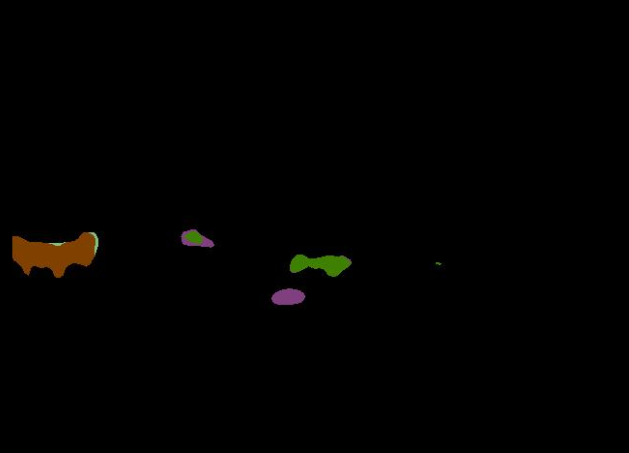}
		\end{subfigure} 
 		\begin{subfigure}[t]{0.24\textwidth}\centering
 			\includegraphics[trim=0cm 0.0cm 0cm 0.0cm, clip=true,width=.98\textwidth]{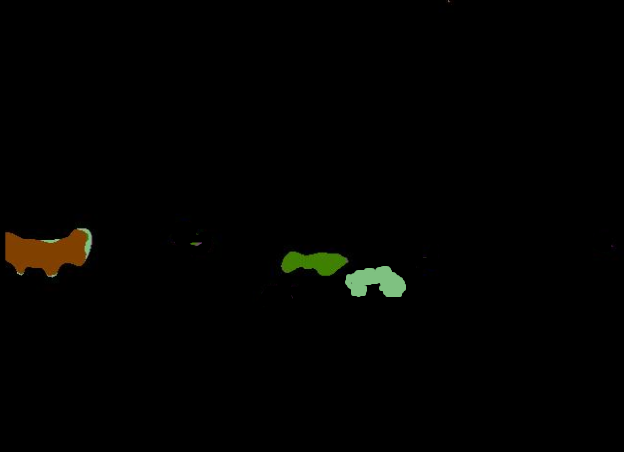}
		\end{subfigure}\\
 	    \begin{subfigure}[t]{0.24\textwidth}\centering 
 			\includegraphics[trim=0cm 0.0cm 0cm 0.0cm, clip=true,width=.98\textwidth]{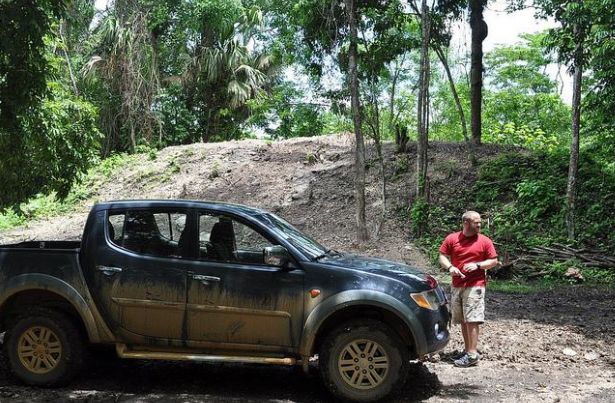}
 		\end{subfigure} 
 		\begin{subfigure}[t]{0.24\textwidth}\centering
 			\includegraphics[trim=0cm 0.0cm 0cm 0.0cm, clip=true,width=.98\textwidth]{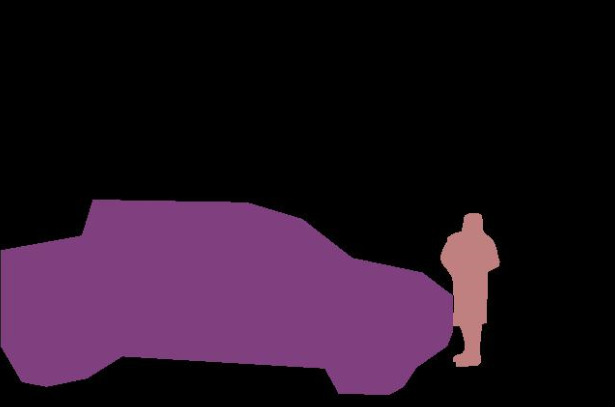}
 		\end{subfigure} 
 		\begin{subfigure}[t]{0.24\textwidth}\centering
 			\includegraphics[trim=0cm 0.0cm 0cm 0.0cm, clip=true,width=.98\textwidth]{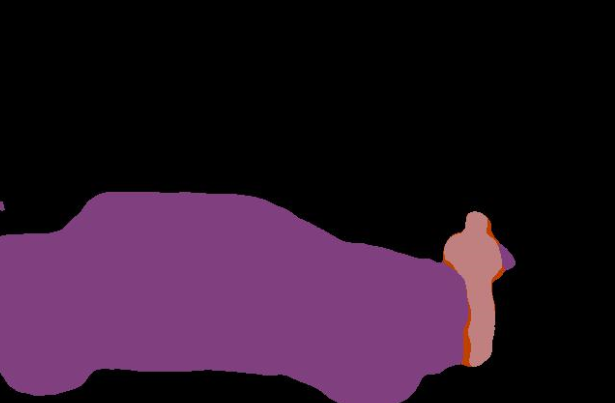}
 		\end{subfigure} 
 		\begin{subfigure}[t]{0.24\textwidth}\centering
 			\includegraphics[trim=0cm 0.0cm 0cm 0.0cm, clip=true,width=.98\textwidth]{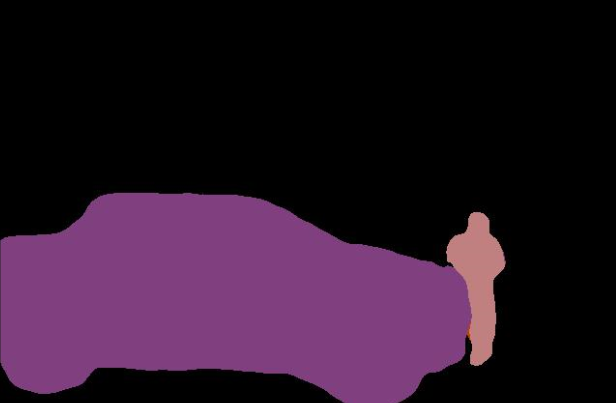}
 		\end{subfigure}
	\end{center}
	\vspace{-0.2cm}
	\caption{\textbf{Qualitative results on MS-COCO}.
	The first and second columns show input images and corresponding segmentation ground truth. The third and fourth columns visualize results produced by ZS3Net and BudaNet.
	Private target classes:
	\setlength{\fboxsep}{1pt}\colorbox{col_bench_coco}{\textcolor{white}{bench}},
	\setlength{\fboxsep}{1pt}\colorbox{col_giraffe_coco}{\textcolor{white}{giraffe}},
	\setlength{\fboxsep}{1pt}\colorbox{col_zebra_coco}{\textcolor{white}{zebra}},
	\setlength{\fboxsep}{1pt}\colorbox{col_truck_coco}{\textcolor{white}{truck}}. 
	Shared classes:
	\setlength{\fboxsep}{1pt}\colorbox{col_person_coco}{\textcolor{white}{person}},
	\setlength{\fboxsep}{1pt}\colorbox{col_cow_coco}{\textcolor{white}{cow}},
	\setlength{\fboxsep}{1pt}\colorbox{col_sheep_coco}{\textcolor{white}{sheep}},
	\setlength{\fboxsep}{1pt}\colorbox{col_background_coco}{\textcolor{white}{background}}.}
	\label{fig:qual_supmat_coco}
	\vspace{-0.2cm}
\end{figure*}
Table~\ref{tbl:cityscapes_res} reports the segmentation performance of the models trained on SYNTHIA data and evaluated on the $19$ classes of the Cityscapes validation set.
Not surprisingly, the pure zero-shot segmentation method with no adaptation (ZS3Net) produces unfavourable results.
The straight-forward baseline (ZS3Net\,+\,UDA) does introduce some improvement over ZS3Net.
With the shared-class alignment strategy proposed in Section~\ref{sec:budanet_shared}, our stronger baseline (ZS3Net\,+\,Adaptation) performs better on all classes.
Finally, BudaNet outperforms all baselines by significant margins for both shared and private classes.
Figure \ref{fig:qual_supmat_cityscapes} illustrates the merit of our approach.
On shared classes, ZS3Net produces noisy predictions, mistaking `road' pixels with `sidewalk', while BudaNet produces correct predictions for most pixels.
On private classes, \textit{i.e.} `train' and `truck', while ZS3Net can only localize small regions, BudaNet provides more complete areas with better contours.
ZS3Net produces noisy segmentation with, for example, the shared class `person' missing in rows 2 and 3. In the same way, the areas of the private class `truck' are hardly detected and wrongly localized elsewhere in rows 1 and 3. Our BudaNet provides better predictions on these areas.

\paragraph{Country-to-Country}

Table~\ref{tbl:idd_res} reports the results on the 21 classes of the IDD validation
set. 
We observe similar behaviors for the baselines and for BudaNet.
While ZS3Net trained only on the source domain produces poor results, ZS3Net\,+\,UDA baseline performs better thanks to the straight-forward domain alignment.
Using our adaptation strategy in Step 1, the ZS3\,+\,Adaptation baseline enhances recognition performances on all classes.
BudaNet introduces gains in both shared and private classes with a significant improvement of $+11.1\%$ hIoU in comparison to ZS3\,+\,Adaptation baseline.
Figure~\ref{fig:qual_supmat_idd} provides some segmentation examples.
Qualitatively, BudaNet's results look much better.
Both ZS3Net and BudaNet produce reasonable segmentation for the shared classes.
However, in rows 1 to 5, ZS3Net hallucinates `auto rickshaw' parts on all vehicles.
Further, most of `auto rickshaw' pixels are partially predicted or incorrectly classified as `truck' (row 3) or `bus' (row 1, 2, 3 and 4).
The last row shows the second private class `animal'.
While ZS3Net completely misses the animals on the road, BudaNet correctly predicts two of them.

\paragraph{Dataset-to-Dataset} 

Table~\ref{tbl:coco_res} reports the semantic segmentation performances on the MS-COCO validation set.
We only report results for the 20 shared\,+\,5 private classes and do not consider others.
Unlike the two other settings, Pascal-VOC and MS-COCO exhibit a small domain gap since they were both collected from the Internet.
Effectively with ZS3Net (no adaptation), the mIoU drop on shared classes is only $4.7\%$ compared to the Oracle.
Still, we demonstrate here similar improvements after addressing adaptation and zero-shot challenges.
Private-class segmentation results remain much lower due to the absence of training samples for these categories.

Domain alignment with the proposed strategy in Step~$1$ (ZS3Net\,+\,Adaptation) brings a $+3.6\%$ mIoU improvement on private classes; the scores on shared classes remain comparable ($63.3\%$ \textit{vs.} $63.8\%$ mIoU).
BudaNet helps boosting the performance in all classes with a hIoU increase of $+7.6\%$ in comparison to ZS3Net, most of which accounts to private classes.
Figure~\ref{fig:qual_supmat_coco} shows some outputs of ZS3Net and BudaNet.
ZS3Net wrongly classifies private objects' parts as background or shared classes.
BudaNet produces better predictions with accurate object contours.
In row 1, while ZS3Net wrongly segments most pixels of the private class `bench' as `chair', BudaNet can recognize more complete objects.
In the 2nd and 3rd examples, ZS3Net confuses the pixels of private `giraffe' and `zebra' classes as `background'.
By contrast, BudaNet delivers good predictions for these classes.
In the last row, BudaNet is shown to improve performance for the `truck' class.

\begin{table*}[h!]
    	\caption{\textbf{Ablation experiments}. Performance in hIoU on the validation set of Cityscapes for five variants of the approach, adding one key component at a time on top of the adaptation-free zero-shot semantic segmentation. See text for details.}
	\begin{center}
		\begin{tabular}{lccccccc} \toprule
			\multirow{2}{*}{Setup} && \multirow{2}{*}{MinEnt} & \multirow{2}{*}{\begin{tabular}[c]{@{}c@{}}Domain Adaptation\\ on shared cls.\end{tabular}} & \multirow{2}{*}{\begin{tabular}[c]{@{}c@{}}Domain-aware ZSL\\ on private cls.\end{tabular}} & \multirow{2}{*}{Self-training} && \multirow{2}{*}{hIoU (\%)}\\\\
			\midrule
			ZS3Net && & & & && 11.1 \\
			ZS3Net\,+\,UDA && \checkmark & & & && 12.8 \\
			BudaNet - Step 1 && \checkmark & \checkmark & & && 14.2 \\
			BudaNet - Step 1+2 && \checkmark & \checkmark & \checkmark & && 22.0 \\
			BudaNet - Step 1+2+3 && \checkmark & \checkmark & \checkmark & \checkmark && \textBF{23.1}\\
			\bottomrule
		\end{tabular}
	\end{center}
	\label{tbl:abl_buda}
\end{table*}
\paragraph{Transductive domain adaptation}
While the focus of the present work is on inductive learning (test data is unknown at train time), transductive learning could also be considered with the proposed tools. In that very specific setup, there is no distinction between train and test data. BudaNet could thus be further refined using pseudo-labeling of the target-domain test set in Step 3. This actually boosts the performance on all classes, with a large gain of $24.9\%$ in harmonic IoU on MS-COCO. Note however that such a transductive scenario is extremely contrived and is not suited for systems to be deployed in open environments.

\begin{table*}[h!]
\centering
\caption{\textbf{Analysis of the adaptation strategy on shared classes.} Target-domain test performance in mIoU on shared (``Shar.'') and private (``Priv.'') classes and in hIOU on both sets (``All'') for the three BUDA settings, and for different adaptation strategies in Step~1.
We compare the proposed stategies ``ZS3Net\,+\,UDA'' and ``ZS3Net\,+\,Adaptation'' to the ZS3Net variant adopting the adversarial DA technique AdvEnt~\cite{vu2019advent}.
Note that ``ZS3Net\,+\,Adaptation'' amounts to ''BudaNet - Step 1'' in Table~\protect{\ref{tbl:abl_buda}}.}
\begin{tabular}{llcccccccccccc} \toprule
 & \multicolumn{3}{c}{Cityscapes} && \multicolumn{3}{c}{IDD} && \multicolumn{3}{c}{MS-COCO}     \\
 \cline{2-4} \cline{6-8} \cline{10-12}\noalign{\smallskip}
Method               & Shar.    & Priv.    & All  && Shar.    & Priv.    & All    && Shar.    & Priv.    & All \\ \midrule 
 ZS3Net\,+\,AdvEnt\,\,\,\,\,\,\,\,\,\,\,\,\,\,   & 30.1 &  7.5 & 12.0 && 32.3 & 8.0 & 12.8 &&  63.4 & 19.5 &  29.8   \\
 ZS3Net\,+\,UDA   & 30.0 &  8.2 & 12.8 && 32.4 & 8.1 & 13.0 &&  63.3 & 19.5 & 29.8 \\
 ZS3Net\,+\,Adaptation  & \textBF{35.0} & \textBF{8.9}  & \textBF{14.2} && \textBF{32.7} & \textBF{8.6} & \textBF{13.6} && \textBF{63.8} & \textBF{20.9} &  \textBF{31.5} \\
 \bottomrule
\end{tabular}
\label{tbl:vanilla_adaptation}
\end{table*}

\subsection{Ablation studies}
\label{sec:ablation}

\paragraph{Effect of the proposed strategies}  
We report in Table \ref{tbl:abl_buda} an ablative study that analyses the impact of four key ingredients of BudaNet: Entropy minimization while training the segmenter $F$ (``MinEnt''); Entropy minimization on top-confident pixels only (``Domain Adaptation on shared classes" 
);  Generator training with adaptation  (``Domain-aware ZSL on private classes" 
); fine-tuning of the semantic segmentation model with pseudo-labels (``Self-training"). To this end, five different models are compared: A model trained only on source, hence without adaptation (``ZS3Net''); Our baseline with entropy minimization (``ZS3Net\,+\,UDA''); The full-fledged model (``BudaNet - Step 1+2+3''); Two stripped-down versions of it (``BudaNet - Step 1'' and ``BudaNet - Step 1+2''). 
We first observe that the brute force entropy minimization contributes to improving the performance by $1.7\%$ (``ZS3Net" \textit{vs.} ``ZS3Net\,+\,UDA"), which demonstrates the need for domain alignment prior to generative model training. The table indicates that the pseudo-label strategy selection for shared-class pixels in the target domain (``ZS3Net\,+\,UDA" \textit{vs.} ``BudaNet - Step 1") significantly increases the performance by $1.4\%$. For the training of the generator, the combination of source-domain training data with pseudo-labeled target-domain data and the adversarial loss (``BudaNet - Step 1" \textit{vs.} ``BudaNet - Step 1+2") further improves the performance, from 14.2 to 22.0. Finally, the complete BudaNet benefits from self-training (up to $1.1\%$), as this step of pseudo-labeling makes it possible to exploit the information related to the private-class visual features.

\begin{figure*}[h!]
    \captionsetup[subfigure]{labelformat=empty}
	\begin{center}
		\begin{subfigure}[t]{0.32\textwidth}\centering
			\caption{(a) Shared}\vspace{-0.2cm}
		\includegraphics[trim=0.45cm 0cm 1.5cm 1.3cm,clip=true,width=1.\textwidth]{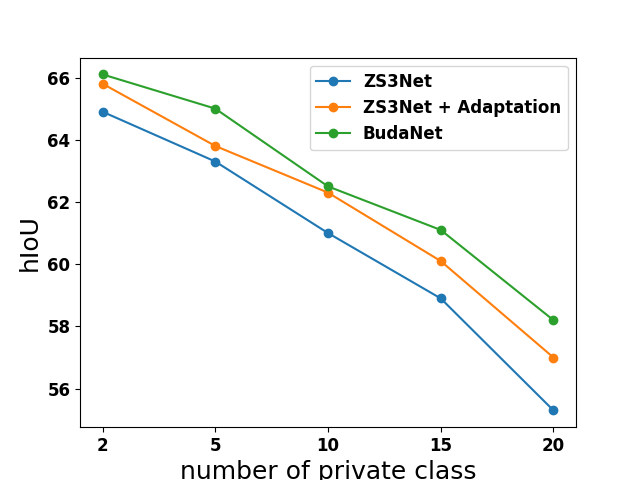}
		\end{subfigure}
		\begin{subfigure}[t]{0.32\textwidth}\centering
			\caption{(b) Private}\vspace{-0.2cm}
				\includegraphics[trim=0.45cm 0cm 1.5cm 1.3cm,clip=true,width=1.\textwidth]{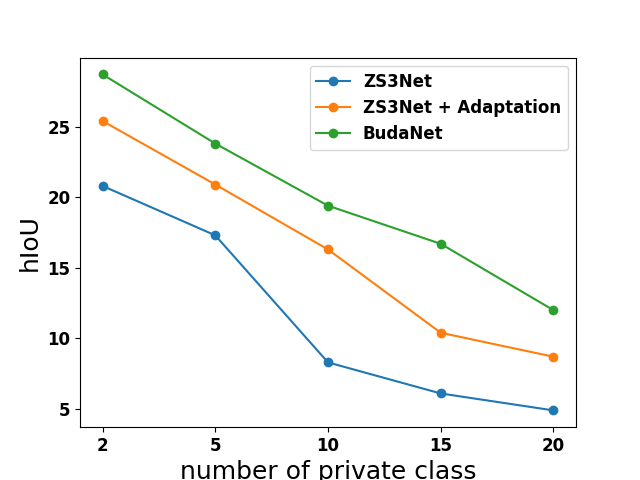}
		\end{subfigure}
		\begin{subfigure}[t]{0.32\textwidth}\centering
			\caption{(c) Overall}\vspace{-0.2cm}
				\includegraphics[trim=0.45cm 0cm 1.5cm 1.3cm,clip=true,width=1.\textwidth]{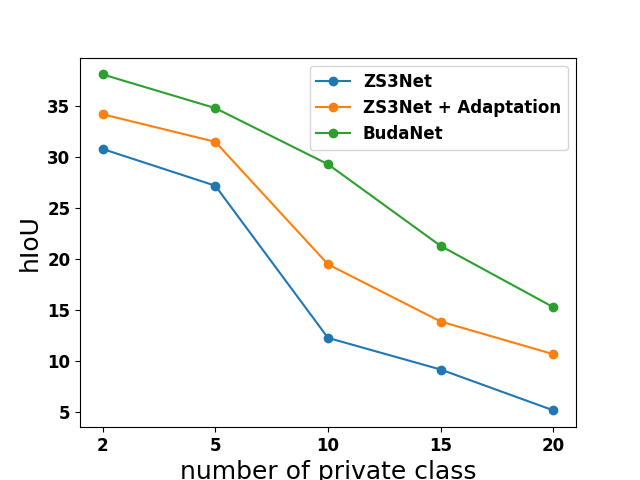}
		\end{subfigure}\\
	\end{center}
	    \vspace{-0.5cm}\caption{\textbf{Influence of the number of target private classes}. Performance in mIoU on shared classes (a), mIoU on private classes (b) and hIoU on both (c) when training BudaNet on Pascal-VOC$\rightarrow$MS-COCO with private class set $\mathcal{C}_p^{2}$, $\mathcal{C}_p^{5}$, $\mathcal{C}_p^{10}$, $\mathcal{C}_p^{15}$ and $\mathcal{C}_p^{20}$, respectively. The number $C_s$ of shared classes is 23, 20, 15, 10 and 5 respectively.}
	\label{fig:nb_private_class} 
\end{figure*}

\begin{figure}[h!]
\centering
\begin{minipage}{0.01\textwidth}
\textbf{\scriptsize \rotatebox{90}{`car'}}
\end{minipage}
\begin{minipage}{0.45\textwidth}
\fcolorbox{white}{gray!8}{
       \includegraphics[scale=0.32, trim={1.6cm 5cm 1.5cm 0.5cm},clip]{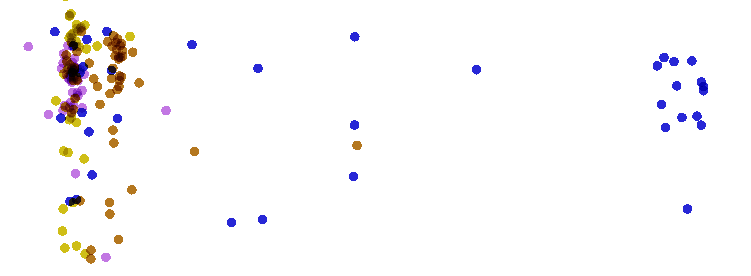}
     }
 \end{minipage}  
 
 \begin{minipage}{0.01\textwidth} 
\textbf{\scriptsize \rotatebox{90}{`tuk-tuk'}}
\end{minipage}
\begin{minipage}{0.45\textwidth}
\fcolorbox{white}{gray!8}{
\includegraphics[scale=0.284, trim={0cm 4.6cm 0cm 3.6cm},clip]{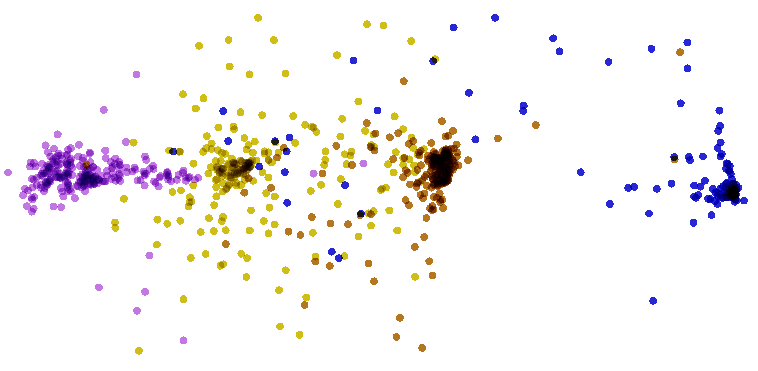}}
 \end{minipage}  
\caption{
\textbf{Domain shifts in visual-feature distributions for shared and private classes}. Visualization with tSNE, for `car' (shared) and `tuk-tuk' (private) classes, of the distributions of (from left to right) \setlength{\fboxsep}{1pt}\colorbox{real_feat}{\textcolor{white}{Real}}, \setlength{\fboxsep}{1pt}\colorbox{buda_feat}{\textcolor{black}{BudaNet}}, \setlength{\fboxsep}{1pt}\colorbox{baseline_feat}{\textcolor{white}{ZS3Net\,+\,UDA}} and \setlength{\fboxsep}{1pt}\colorbox{zs3_feat}{\textcolor{white}{ZS3Net}} features. Must be viewed in color.}
\label{fig:visu_gen}
\vspace{-0.6cm}
\end{figure}

\paragraph{Number of private classes}
To analyse the impact of the number of private classes on the performance of BudaNet, we made it vary in the \emph{dataset-2-dataset} setup.
More precisely, we defined five nested sets with 2, 5, 10, 15 and 20 private classes respectively, among the 25 classes in Pascal-VOC and MS-COO. These sets are:  
\begin{align*}
    \mathcal{C}_{p}^2 &= \{\text{`truck',`zebra'}\},\\
    \mathcal{C}_{p}^5 &= \mathcal{C}_{p}^{2} \cup \{\text{`bench',`giraffe',`laptop'}\},\\
    \mathcal{C}_{p}^{10} &= \mathcal{C}_{p}^{5} \cup \{\text{`elephant',`umbrella',`bear',`snowboard',`toilet'}\},\\
    \mathcal{C}_{p}^{15} &= \mathcal{C}_{p}^{10} \cup \{\text{`laptop',`fridge',`blanket',`napkin',`stone'}\},\\
    \mathcal{C}_{p}^{20} &= \mathcal{C}_{p}^{15} \cup \{\text{`tent',`skyscraper',`salad',`river',`pillow'}\}.
\end{align*}
The corresponding sets of shared labels have size 23, 20, 15, 10 and 5 respectively. Note that $\mathcal{C}_{p}^5$ is the private label set used in the main dataset-to-dataset experiments.

Figure \ref{fig:nb_private_class} shows hIoU plots as a function of the number of private classes.
We observe a decreasing trend in performance for all methods when more private classes (respectively fewer shared classes) are used.
In all set-ups, BudaNet (green curves) outperforms the baselines by a significant margin.

\paragraph{Domain adaptation strategy on shared classes}
In Table~\ref{tbl:vanilla_adaptation}, we compare the proposed DA strategies on shared classes to the state-of-the-art adversarial DA technique AdvEnt~\cite{vu2019advent}.
We here notice the negative effects caused by global alignment techniques like AdvEnt in BUDA; such a drawback was similarly observed in partial DA and open-set DA works.

These experiments echo the arguments developed in Section~\ref{sec:budanet_shared} to justify the relevance of the proposed adaptation strategy.

\paragraph{Feature visualization}
Using the models in Table~\ref{tbl:idd_res}, we visualize in Figure~\ref{fig:visu_gen} domain shifts between real and generated pixel-wise features from different approaches for both shared and private classes.
For the shared class `car', as real features are used in training, we observe equally good alignment from BudaNet and ``ZS3Net\,+\,UDA'', as opposed to the adaptation-free ZS3Net.
Regarding the private class `auto-rickshaw', we clearly see a smaller gap between real and BudaNet-generated features, compared to the baseline and to the adaptation-free model.
As no real private-class features are provided during training, we do not expect a ``perfect'' alignment in the latter case.
The visualization is inline with the results in Table~\ref{tbl:idd_res}.

\begin{table}[h!]
\caption{\textbf{Impact of the fraction of high-scoring predictions retained as pseudo-labels}. Target-domain performance in hIoU on the SYNTHIA$\rightarrow$Cityscapes setup as a function of $p(\%)$. ``GT": see text for explanation.} 
    \vspace{-.3cm}
    \small
	\begin{center}
		\setlength{\tabcolsep}{3.5pt} 
		\begin{tabular}{l|ccccc|c} \toprule
			\rule{0pt}{3ex} $p\%$ of high-scoring &10\%&30\%&50\%&70\%&100\%& GT \\
		\midrule
			\rule{0pt}{3ex}Cityscapes hIoU & 20.4  & 21.6 & \textbf{22.0} & 21.8 & 17.8 & 38.9 \\
			\bottomrule
		\end{tabular}
	\end{center}
	\label{tbl:abl_percentage_top}
\end{table}
\paragraph{Proportion of retained top-scoring predictions as pseudo-labels in Step 2}
Our aim here is to compare different $p\%$ values for the pseudo-labeling strategy on target features. Table \ref{tbl:abl_percentage_top} reports hIoU performance on Cityscapes as this parameter varies. Keeping only the highest-scoring predictions, at $p=10\%$, results in a drop of performance of $1.6\%$. One possible explanation is that because the number of training features becomes too low, the model lacks generalization capacity. Conversely, when all predictions are kept ($p=100\%$), a significant decrease of performance can be observed, which confirms the need for an effective pseudo-label selection strategy. Finally, we observe that the performance is better when $p\in \left [ 30, 70 \right ]$, specifically when $p=50\%$ with a recognition score of $22.0$ hIoU~\footnote{The result corresponds to ``BudaNet - Step 1+2'' in Table~\ref{tbl:abl_buda}.}.
In the last column of Table \ref{tbl:abl_percentage_top}, ``GT'', we report the recognition score when the $p=50\%$ highest-scoring predictions are replaced by their ground-truth labels. We observe a performance improvement of $16.9\%$, the pixel selection technique is therefore an essential element of the method and must be investigated in future work.
In all other steps with pseudo-labeling, we also fix the percentage of retained high-scoring predictions as $50\%$.

\section{Conclusion}
In the context of semantic segmentation, we have explored boundless UDA, the realistic domain adaptation setting where new classes can emerge in the target test domain.
Compared to existing DA settings, BUDA is more challenging because (i) objects from private new classes must be explicitly recognized at test time and (ii) objects from both shared and private classes frequently coexist in the test scenes.
To address this new problem, we proposed BudaNet, a unified pipeline that leverages domain adaptation and zero-shot learning techniques.
BudaNet consists of three steps with dedicated strategies to (i) mitigate the domain gap, (ii) handle the private classes and (iii) adjust the decision boundaries in the presence of new-class objects.
On generalized ZSL evaluation metrics, BudaNet outperforms the baselines by significant margins and sets competitive standard in the BUDA setting.
This first step towards boundless UDA will hopefully foster future research on this type of domain adaptation and, more generally, on new DA settings motivated by real-world applications.

\clearpage

\bibliographystyle{elsarticle-num}
\bibliography{0-bibfile}

\end{document}